\documentclass[10pt,twocolumn,letterpaper]{article}

\usepackage{iccv}
\usepackage{times}
\usepackage{epsfig}
\usepackage{graphicx}
\usepackage{amsmath}
\usepackage{amssymb}
\usepackage{breqn}
\usepackage{float}
\usepackage{algorithm}
\usepackage{algorithmic}
\usepackage{amsmath}
\usepackage{capt-of}
\usepackage{pdfpages}
\usepackage{enumerate}
\usepackage{footnote}
\usepackage{multirow,xcolor}
\usepackage{makecell}
\usepackage{subcaption}
\usepackage[pagebackref=true,breaklinks=true,letterpaper=true,colorlinks,
citecolor=black,bookmarks=false]{hyperref}

\iccvfinalcopy % *** Uncomment this line for the final submission

\def\iccvPaperID{782} % *** Enter the ICCV Paper ID here

\ificcvfinal\fi

\begin{document}

%%%%%%%%% TITLE
\title{Make a Face: Towards Arbitrary High Fidelity Face Manipulation}

\author{Shengju Qian\textsuperscript{1}\footnotemark~~~  Kwan-Yee Lin\textsuperscript{5}~~~ Wayne Wu\textsuperscript{2,5}~~~ Yangxiaokang Liu\textsuperscript{3}~~~ Quan Wang\textsuperscript{5} \\~~~ Fumin Shen\textsuperscript{3}~~~ Chen Qian\textsuperscript{5}~~~ Ran He\textsuperscript{4} \\
\textsuperscript{1}The Chinese University of Hong Kong~~~ \textsuperscript{2}Tsinghua University \\ \textsuperscript{3}University of Electronic Science and Technology of China~~~ \textsuperscript{4}NLPR, CASIA  \\
\textsuperscript{5}SenseTime Research
}

\twocolumn[{
\renewcommand\twocolumn[1][]{#1}
\maketitle
\begin{center}
\vspace{-0.5cm}
    \centering
    \includegraphics[width=0.85\textwidth]{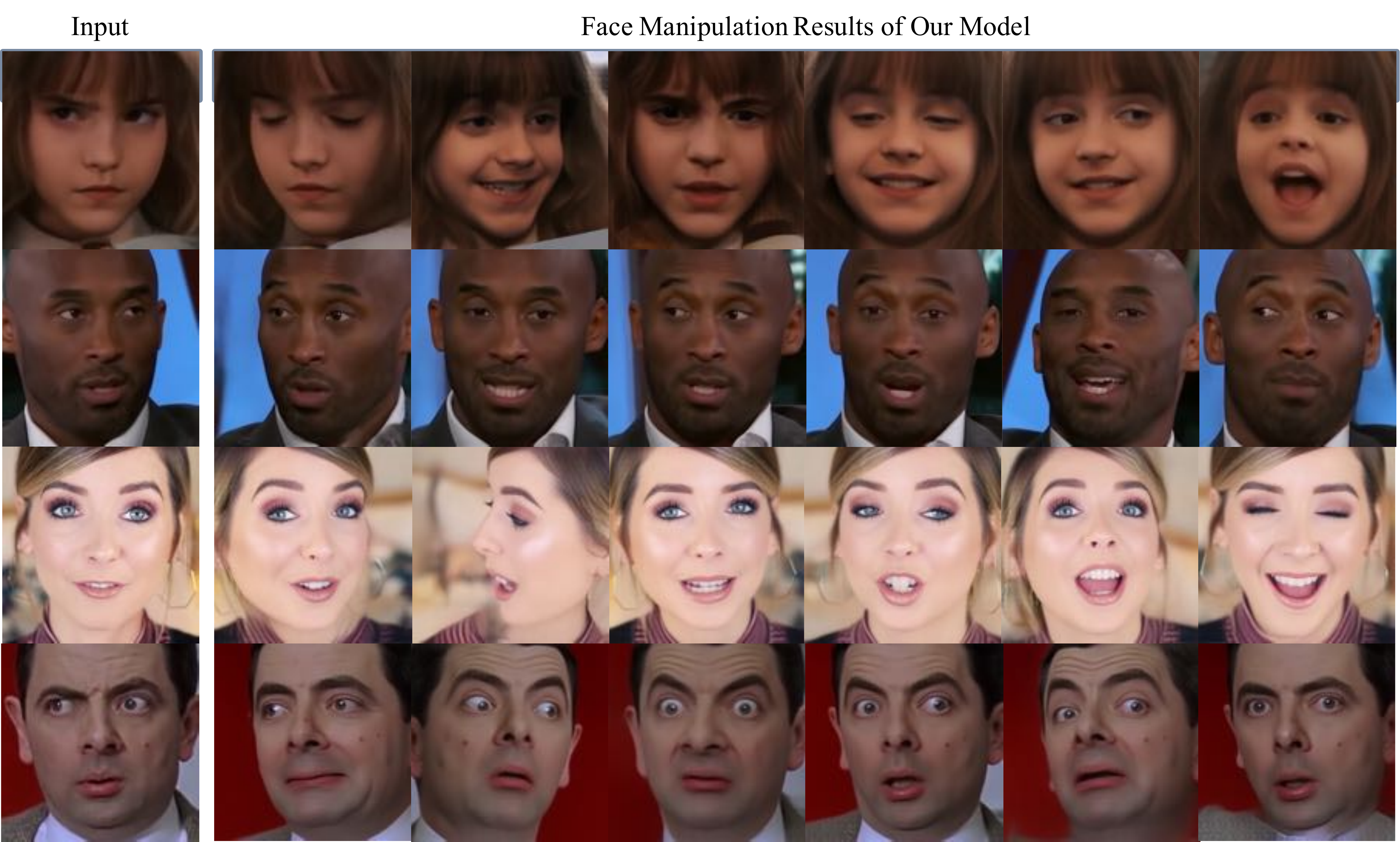}
    \captionof{figure}{\small{Face manipulation results on in-the-wild samples via transferring knowledge learned from the CelebA dataset. The first column shows input images and the remainders are images generated by AF-VAE with target expression/rotation boundary maps as the condition. Note that the model is fine-tuned with movie clip frames from YouTube of  $256\times256$ resolution. All the generated poses are unseen before.}}
    \label{inthewild}
\end{center}
}]

\vspace{-1cm}

\maketitle
  \renewcommand*{\footnote}{\fnsymbol{footnote}}
  \setcounter{footnote}{1}
  \footnotetext{Work done during an internship at SenseTime Research.}
  \renewcommand*{\footnote}{\arabic{footnote}}

\thispagestyle{empty}
%%%%%%%%% ABSTRACT

\begin{abstract}
 Recent studies have shown remarkable success in face manipulation task with the advance of GANs and VAEs paradigms, but the outputs are sometimes limited to low-resolution and lack of diversity.

 In this work, we propose Additive Focal Variational Auto-encoder (AF-VAE), a novel approach that can arbitrarily manipulate high-resolution face images using a simple yet effective model and only weak supervision of reconstruction and KL divergence losses. First, a novel additive Gaussian Mixture assumption is introduced with an unsupervised clustering mechanism in the structural latent space, which endows better disentanglement and boosts multi-modal representation with external memory. Second, to improve the perceptual quality of synthesized results, two simple strategies in architecture design are further tailored and discussed on the behavior of Human Visual System (HVS) for the first time, allowing for fine control over the model complexity and sample quality. Human opinion studies and new state-of-the-art  Inception Score (IS) / Fr\'echet Inception Distance (FID) demonstrate the superiority of our approach over existing algorithms, advancing both the fidelity and extremity of face manipulation task.
\end{abstract}

%%%%%%%%% BODY TEXTi c
\section{Introduction}
Automatically manipulating facial expressions and head poses from a single image is a challenging open-end conditional generation task. Faithful photo-realistic face manipulation finds a wide range of applications in industry, such as film production, face analysis, and photography technologies. With the flourish of generative models, the state of this task has advanced dramatically in recent years at the forefront of efforts to generating diverse and photo-realistic results. Nevertheless, it is highly challenging for a generative model to learn a compact representation of intrinsic face properties to synthesize face images with \textit{high-fidelity} or \textit{large facial expression/poses}, due to the ill-posed nature of lacking paired training data.

Current state-of-the-art face manipulation approaches
~\cite{DBLP:conf/eccv/PumarolaAMSM18,Kossaifi_2018_CVPR,Choi_2018_CVPR,Shen_2018_CVPR} mainly benefit from the advancement of generative adversarial networks (GANs). To tackle the two bottlenecks mentioned above, vast algorithms focus on sophisticated modifications to loss term or generator architecture, with the injection of different facial attribute information ~\cite{yang2018pose,Shen_2018_CVPR,Kossaifi_2018_CVPR,wang2018every,wiles2018x2face,qiao2018geometry,Wiles18a,lu2018attribute}. Other works focus on the design of task-specific training procedure~\cite{yang2018pose,hu2018pose,wang2018every,towards-large-pose-face-frontalization-in-the-wild,chen2018facelet,Bao_2018_CVPR,ma2017pose,ma2018disentangled}. Nevertheless, successful generation of plausible samples on extreme face geomorphing and complex uncontrolled datasets remain elusive goals for these methods due to their unstable training procedures and environmental constraints. The current state-of-the-art~\cite{DBLP:conf/eccv/PumarolaAMSM18} in RaFD~\cite{liu2015faceattributes} face expression synthesis achieves a Fr\'echet Inception Distance (FID)~\cite{heusel2017gans} of $34$, still leaving a large gap towards real data even in a controlled environment.

In this work, we set out to close the gap in fidelity and extremity between facial expressions/rotations generated by current state-of-the-arts and real-world face images with a simple yet effective framework. We explore the conditional variational auto-encoder (C-VAE) formalism~\cite{Kingma2014,sohn2015learning} for face manipulation task. It is intuitive to adopt C-VAE by taking advantage of its nice manifold representation and stable training mechanism. Nonetheless, tailoring the ~``vanilla'' C-VAE to face manipulation task is \textit{non-trivial}: $(1)$ The diversity of synthesized outputs will be sacrificed since the latent distribution is commonly assumed to be a unit Gaussian. However, as visualized in Fig.~\ref{framework}~(b), a complex latent representation that needs to describe factors like ages, complexions, luminance, and poses would break the Gaussian assumption due to its insufficiency. $(2)$ Facial expressions are more fine-grained than other sources like landscapes, digits, and animals. Thus the property and common architecture of VAE could not satisfy the requisites for maintaining facial details on high-resolution images.

As a solution to such problems, we propose a novel Additive focal variational auto-encoder (AF-VAE) framework. By applying a light-weight geometry-guidance to explicitly disentangle facial appearance and structure in latent space, we encourage the latent code to be separated into a pose-invariant appearance representation and a structure representation, thereby preserving the appearance and structure information under geomorphing. To tackle the issue of diversity, a novel additive memory module that bridges unsupervised clustering mechanism with Gaussian Mixture prior in the structured latent space is introduced to the framework, endowing the ability of multi-model facial expression/rotations generation.

To further improve the perceptual quality of synthesized results, we also discover two simple yet effective strategies in model design and characterize them empirically with the behavior of Human Visual System (HVS). Leveraging insights from this empirical analysis, we demonstrate that a simple injection of these strategies can easily improve the perceptual quality of synthesized results. Our model can steadily manipulate photo-realistic facial expression and face rotations at $256\times256$ resolution under uncontrolled settings. The proposed AM-VAE improves the state-of-the-art Fr\'echet Inception Distance (FID) and Inception Score (IS) on uncontrolled CelebA~\cite{liu2015faceattributes} from $71.3$ and $1.065$ to $36.82$ and $2.15$.

 We conduct comparisons with current state-of-the-art face manipulation algorithms~\cite{Choi_2018_CVPR,DBLP:conf/eccv/PumarolaAMSM18,Shen_2018_CVPR,wang2018vid2vid} and show that our approach outperforms these methods regarding both quantitative and qualitative evaluation. Extensive self-evaluation experiments further demonstrate the effectiveness of proposed components.

\section{Related Work}
\noindent
\textbf{Face Manipulation.} In the literature of face image manipulation, besides classic mass-spring models and 2D/3D morphing methods~\cite{thies2016face2face,nonlinear-3d-face-morphable-model,on-learning-3d-face-morphable-model-from-in-the-wild-images}, recently, significant progress has been achieved by leveraging the power of Generative Adversarial Networks (GANs)~\cite{DBLP:conf/eccv/PumarolaAMSM18,Kossaifi_2018_CVPR,Choi_2018_CVPR,Shen_2018_CVPR,wayne2018reenactgan,wayne2019transgaga} for photo-realistic synthesis results. To improve the robustness and diversity of GANs, tweaks on various aspects are explored. For example, StarGAN~\cite{Choi_2018_CVPR} exploits cycle consistency to preserve key attributes between the source and target images. GANimation~\cite{DBLP:conf/eccv/PumarolaAMSM18} takes a step further, utilizing denser AUs prior vector to increase the diversity by altering the magnitude of each AUs. Attention mechanism in generator architecture is also used to mask out irrelevant facial regions, enforcing the network to only synthesize in-region texture. FaceID-GAN~\cite{Shen_2018_CVPR} introduces a three-player adversarial scheme and utilizes 3DMM~\cite{blanz1999morphable} to better preserve facial property. CAPG-GAN~\cite{hu2018pose} trains a couple-agent discriminator to constrain both distributions of pose and facial structure.

Since GANs face challenges in brittle training procedure and sampling diversity, some works take advantages from variational auto-encoder (VAE) paradigm and its variants with an exploration of disentangling intrinsic facial properties in latent space. For example,
Neural Face Editing~\cite{DBLP:conf/cvpr/ShuYHSSS17} and Deforming Autoencoders~\cite{DBLP:conf/eccv/ShuSGSPK18} utilize graphics rendering elements such as UV maps, albedo, and shading to decouple the latent representation. However, the main disadvantages of these VAE-based methods are the blurry synthesized results caused by injected element-wise divergence measurement and imperfect network architecture.  CVAE-GAN~\cite{DBLP:conf/iccv/BaoCWLH17} combines VAE and GAN into a framework with an asymmetric training loss and fine-grained category label. These methods lack an interface for users to manipulate facial expression arbitrarily since they edit the results through manifold traversal. Notably, the capacity constraint of VAE is also discussed in other tasks like image caption, in~\cite{wang2017diverse,qianextending}, an additive Gaussian encoding space is proposed to provide a more diverse and accurate caption result. Motivated by those prior works, we exploit AF-VAE, which can compensate for the drawbacks of VAEs by providing a conditional geometry-related additive memory prior as well as two light-weight network design strategies to the framework.

\noindent
\textbf{High Fidelity Image Synthesis.}
Recently, generating high-resolution samples with fine details and realistic textures has become a trend in image synthesis task. For instance, pix2pixHD~\cite{Wang_2018_CVPR} introduces a coarse-to-fine generator, a multi-scale discriminator and a feature matching loss on the basis of pix2pix~\cite{DBLP:conf/cvpr/IsolaZZE17}. It could translate facial edge to photo-realistic face image under $1024\times1024$ resolution. While it requires paired training data and thereby cannot be generalized to arbitrary edges. ~\cite{DBLP:journals/corr/abs-1801-07632} utilizes progressive GANs to generate $512\times512$ face images under the control of discrete one-hot attributes. IntroVAE~\cite{huang2018introvae} proposes to jointly train its inference and generator in an introspective way, achieving $1024\times1024$ resolution on reconstruction. BigGAN~\cite{brock2018large} modifies a regularization scheme and a sampling technique to class-conditional GANs to achieve $512\time512$ resolution on ImageNet. While all these frameworks are only capable of uncontrolled generation or discrete attribute editing, our framework focuses on challenging high fidelity face manipulation without leveraging paired training data.

\section{Method}{\label{method}}
In this section, we explore methods for manipulating facial expression/rotation and scaling up modal training to reap the benefits of detailed architecture design. We first clarify our notations here. Given a face image $x$ from a dataset $X$, the goal of our task is to learn a mapping $\mathcal{G}$ to transfer $x$ to an output image $\tilde{x}$, conditioned on a target facial structural information $c$. The intrinsic challenge lying behind this task is to preserve facial \textit{appearance} under the high-fidelity setting and dramatic \textit{structural} changes. In this light, we decouple the mapping $\mathcal{G}$ into $\phi_{app}$ and $\mu_{str}$, which are expected to learn the pose-invariant appearance representation $z=\phi_{app}(x,c)$  and rational structure representation  $y=\mu_{str}(c)$ respectively.
% In this section, we explore methods for manipulating face expression/rotations and scaling up model training to reap the performance benefits of light-weight architecture designing. We first clarify our notations here. Given a face image $x$ from a dataset $X$, the goal of our task is to learn a mapping $\mathcal{G}$ to transfer $x$ to an output image $\tilde{x}$, conditioned on a target facial expression or rotation information $c${\footnote{Note that, neither the training pairs $(x,\tilde{x})$ nor the supervision signals that related to $c$ are required under our weakly supervised face manipulation framework.}}. The intrinsical challenges lie in this task are the \textit{appearance preserving} and \textit{structure rationality} problems, when comes to large geometry changes or high fidelity setting. In this light, we decouple the mapping $\mathcal{G}$ into $\phi_{app}$ and $\mu_{str}$, which are expected to learn the pose-invariant appearance representation $z=\phi_{app}(x,c)$  and rational structure representation  $y=\mu_{str}(c)$ respectively.

In this way, the conditional variational auto-encoder (C-VAE) can be tailored as a baseline for its capability of disentangling $z$ from $y$ by maximizing the lower bound on the conditional data-log-likelihood $p(x|y)$, \ie,
\begin{equation}\label{eq1}
\log p(x|y) \ge \mathbb{E}_q[\log p(x|z,y)]-D_{KL}[q(z|x,y),p(z|y)],
\end{equation}
where $q(z|x,y)$ is an approximate distribution of posterior $p(z|y)$. In particular, $q(z|x,y)$ and $p(x|z,y)$ are the \textit{encoder} and \textit{decoder} respectively. The model is typically trained with the following stochastic objective:
  \vspace{-0.2cm}
 \begin{align}\label{eq2}
 \mathcal{L}(x,\phi,\varphi)& = -\frac{1}{N}\sum_{i=1}^N\log p_{\varphi}(x^i|z^i,y^i) + \\ & D_{KL}[q_{\phi}(z|x,y),p(z|y)], \nonumber s.t.\forall i \;z^i\sim q_{\phi}(z|x,y)
 \end{align}
where $\phi$ and $\varphi$, the parameters for the encoder and decoder, are learned with reparameterization trick~\cite{Kingma2014}. $q_\phi(z|x,y)$ is typically restricted to be a distribution over $\mathcal{N}(\mathbf{0},\mathbf{I})$ as in~\cite{Kingma2014}.

% Since the KL term encourages $q(z|x,y)$ to be close to the prior $p(z|y)$, the behavior of $z$ should only rely on $x$, and enforced to be invariant to $y$.  Ideally, if the decoder can perfectly reconstruct the $x$, the first term would be a delta function over $x$. Thus, we could understand the first term of Equation~\ref{eq2} as a reconstruction loss $L_{rec}$, and the second term as a regularizer $L_{KL}$ in auto-encoder parlance.

However, there are three problems to discuss under the formulation of problems:
$(1)$How to accurately disentangle the latent space to guarantee that $z$ is complementary to the structure prior $y$, with nothing but appearance?
$(2)$ The KL divergence term facilitates the structure of the learnt latent space to get close to the prior $p(z|y)$ over $\mathcal{N}(\mathbf{0},\mathbf{I})$. Does this choice satisfy the encoder's modeling of the diversity of the face samples?
$(3)$ Does the perceptual quality of synthesized results look appealing to viewers with only the reconstruction and KL divergence losses?

\begin{figure*}[t!]
\begin{center}
\includegraphics[width=0.9\linewidth, angle=0]{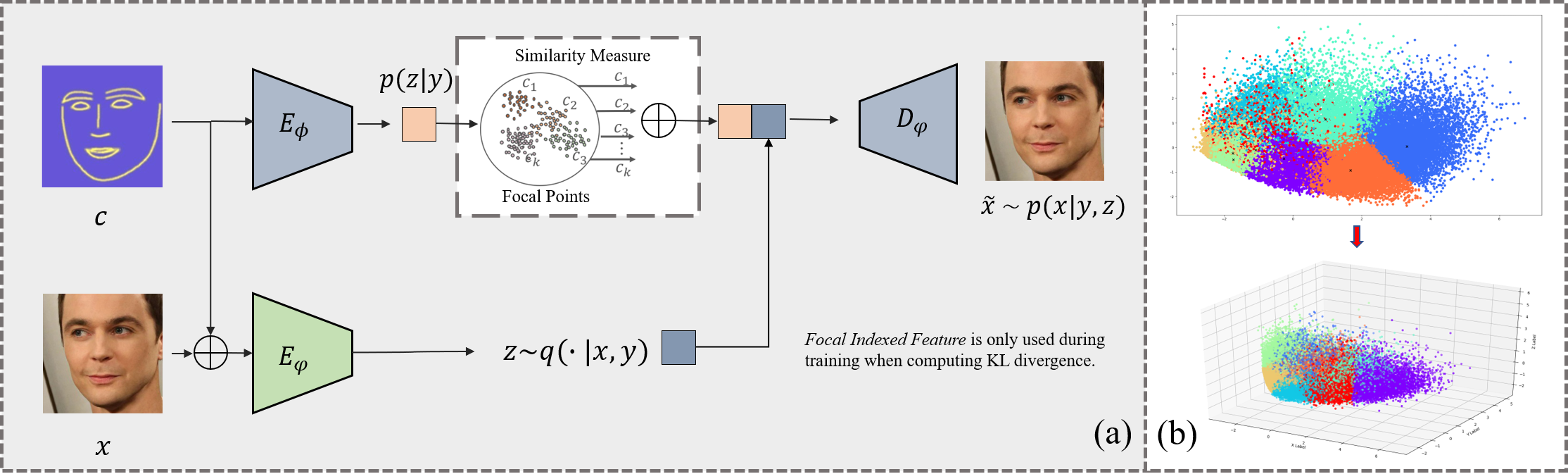}
\end{center}
\vspace{-0.5cm}
\caption{\label{framework}\small{(a)~Our framework (b)~2D and 3D projection of 5000 facial structure representations. Each color denotes a cluster, a more intuitive illustration of our approach is to map each cluster to a Gaussian prior, extending the capacity of C-VAE constrained by single prior.}}
\vspace{-0.6cm}
\end{figure*}

We concern the optimality of these choices on face manipulation and explore better solutions in the following sections. An overview of our framework is illustrated in Fig.~\ref{framework}(a).

\subsection{Geometry-Guided Disentanglement}
In order to arbitrarily manipulate a face image, sufficient geometry information should be provided in $c$. Typical choices of manipulated condition include facial landmarks, 3DMM parameters and masks. On the basis of sparse landmarks information as used in Pix2pixHD~\cite{Wang_2018_CVPR}, we use a offline interpolation process in~\cite{Wu_2018_CVPR} to obtain boundary maps. This off-line process is formulated as $c=F_\omega(x)$. Then the latent structure representation $y$ could be obtained by encoding $c$ with an encoder $E_\mu$.

Without any semantic assumption, the framework can only make sure the latent code $z$ drawn from $x$ is invariant to structure $y$. To this end, we leverage the distilled geometry information $y$ to disentangle the latent space \textit{explicitly}. Concretely, $y$ is concatenated with the inferred appearance representation $z$. Then, the concatenated representation is forwarded to decoder $D_\varphi$. The skip-connections between $E_\mu$ and $D_\varphi$ are further incorporated to potentially ensure enough structure information obtained from the prior (\ie, the decoder $D_\varphi$). Thus, $z$ is encouraged to encode more information about appearance rather than structure, otherwise, a penalty of the likelihood $p(x|y,z)$ will be incurred for large reconstruction error.

\subsection{Additive Memory Encoding}
While the explicit geometry-guided disentanglement helps preserving facial structure and structuring latent appearance representation to be invariant towards geomorphing, it is still hard to meet the requirement of fine-grained face manipulation. As shown in Fig.~\ref{ablation}, synthesized results on profile with extreme expression or under in-the-wild environment could easily collapse into a ~``mean-face'' phenomenon. The C-VAE formulation is restricted to draw the prior $p(z|y)$ from a simple structure, typically a zero-mean unit variance Gaussian. However, factors like ages, complexions, luminance, and poses constitute a complex distribution, which breaks the common assumption. Consequently, some peculiar facial features might become ~``outliers'' of the distribution during training.

To this end, a better choice of the prior structure is to be explored for a decent appearance representation. Intuitively, drawing the complex prior from a combination of simple distributions will increase the diversity of latent code $z$ and meanwhile enable computing the closed form.

We thus encourage the appearance representation $z$ to have a multi-modal structure composed of $K$ clusters, each corresponding to a semantic feature. In practice, we construct the memory bank through K-means clustering on all boundaries in training set. Every cluster center is called a \textit{focal point}, which commonly refers to a distinctive characteristic such as laughing or side face. In this way, each boundary map used in training would have a k-dimension \textit{focal indexed feature}: $w(b)=(w_1(b),w_2(b),\cdots,w_k(b))$, denoting its similarity measurement with each focal point. The focal indexed feature is used in training to boost the diversity and plentiful appearance in the latent representation. Since the external memory contains large geometry variation, the general idea is to build up latent representation capacity leveraging explicit and concise spatial semantic guidance provided by focal points. By modeling $p(z|y)$ as a Gaussian Mixture, for each cluster $k$ with weight $w_k$, mean $\mu_k$ and standard deviation $\sigma_k$, we have:
{\setlength\belowdisplayskip{1pt}
\begin{equation}\label{eq3}
p(z|y)=\sum_{k=1}^K w_k\mathcal{N}(z|\mu_k,\sigma_{k}^2I).
\end{equation}}

However, as it is not directly tractable to optimize Equation~\ref{eq2} with GMM prior, the KL divergence needs to be approximated in training, sampling $z$ from one of the clusters according to their probability. Hence this operation fails to model targets that contain more than one facial geometry characters, \eg, laughing and side face simultaneously. Consequently, we introduce an Additive Focal prior to our framework with the formulation as:
{\setlength\abovedisplayskip{3pt}
\setlength\belowdisplayskip{2pt}
\begin{align}
	p(z\vert y) = N(z\vert\sum_{k=1}^K w_k \mu_k, \sigma^2I)
	\label{eq:probAG}
\end{align}}

where $\sigma^2 I$ is a co-variance matrix with $\sigma^2 = \sum_{k=1}^Kw_k^2 \sigma_k^2$. Behind the formula, we assume the face image contains multiple structural characteristics with weights $w_k$. The value corresponds to similarity measurement with each focal point, which is defined by a normalized cosine distance. The means $\mu$ of clusters are randomly initialized on the unit ball. The KL term could be computed over $q(z\vert x,y) = N(z\vert \mu(x,y),\sigma^2(x,y)I)$ , which can be derived to be:
\vspace{-0.1cm}
\begin{align}
	D_{KL} & = \log (\frac{\sigma}{\sigma_\phi}) + \frac{1}{2\sigma^2}E_{q_\phi}[(z-\sum_{k=1}^K w_k \mu_k)^2] - \frac{1}{2} \nonumber\\
	&= \log (\frac{\sigma}{\sigma_\phi}) + \frac{\sigma_\phi^2 + (\mu_phi - \sum_{k=1}^K w_k \mu_k)^2}{2\sigma^2} - \frac{1}{2}
	\label{eq5}
\end{align}
By combining the above KL term into Equation~\ref{eq2}, we obtain the final loss function to train the generator.

\subsection{Quality-Aware Synthesis}

Given a source image $x$ and target boundary $\tilde{b}$, we are able to manipulate the facial expression and head poses of $x$ through the proposed AF-VAE. Following ~\cite{chen2017photographic,esser2018variational}, we use feature matching loss $L_{rec}= \|x - \hat{x}\|_1 + \sum_l \lambda_l \| \psi_l (x) - \psi_l (\hat{x})\|$ as reconstruction loss to overcome the blurry results of $L_1$/$L_2$ losses. However, as can be seen from Table~\ref{ablation_fid}, the results are still far from being perceptually appealing.

There are many factors that may cause artifacts in synthesized images and result in perceptual quality distortion, such as losses, unstable training process, and network architectures. Intuitively, one can ease the problem by incorporating auxiliary objective functions~\cite{gulrajani2017improved,mao2017least}, or designing sophisticated attention mechanism to capture better global structure~\cite{DBLP:conf/eccv/PumarolaAMSM18}. However, we turn to two light-weight designs in network structure through observations and HVS basis for the first time.

\begin{figure}[]
\begin{center}
\includegraphics[width=0.9\linewidth, angle=0]{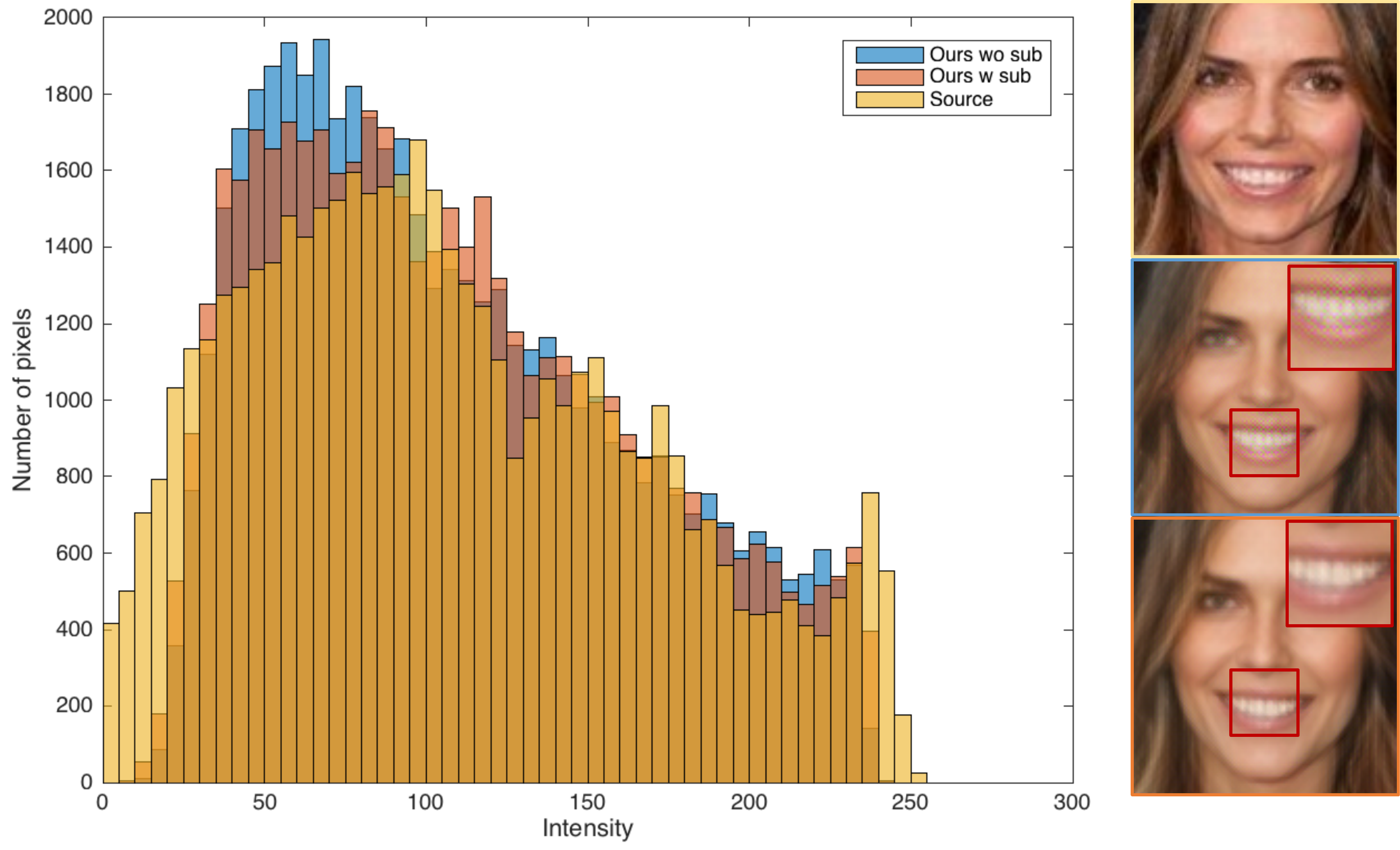}
\end{center}
\vspace{-0.5cm}
\caption{\label{his}\small{Histograms w/wo pixel-shuffle (Better zoom in).}}
\vspace{-0.5cm}
\end{figure}

Instead of using deconvolution operation to perform upsampling in decoder, we use sub-pixel convolution~\cite{shi2016real} in every upsamping layer. As shown in Fig.~\ref{his}, the ~`checkerboard' artifacts are reduced apparently. However, it is interesting that the reconstruction results generated from AF-VAE model with/without subpixel yield similar distributions over their histograms, and the differences of their entropy are small ($7.6405$ without subpixel, $7.6794$ with subpixel, and $7.8267$ for the source image). It means that subpixel does not filter the artifacts substantially, it scatters them into parts of the image instead. However, HVS treats artifact signals unevenly according to areas of images. Artifacts lie in low frequency or edge areas are emphasized and ones lie in high frequency areas are tend to be masked, as demonstrated in the task of perceptual metrics learning~\cite{Pan_2018_CVPR,Lin_2018_CVPR,Liu_2017_ICCV,zhang2018perceptual}. Thus, we can draw a reasonable explanation for subpixel convolution on the improvements of perceptual measurements.

Another technique is that we utilize Weight Normalization(WN)~\cite{salimans2016weight} for model training. It is habitual in GAN-based methods to replace BatchNorm(BN)~\cite{ioffe2015batch} to WN or its variant for stabilizing the training of the \textit{discriminator}. Instead, we tailor this strategy to both \textit{encoder} and \textit{decoder} of AF-VAE. Models trained with weight normalization have faster convergence and lower reconstruction loss than the ones without WN over training iterations. Detailed analysis and loss curve plot are provided in the appendix. This phenomenon is similar to GANs. However, as can be seen from Fig.~\ref{ablation}, WN helps increase the diversity of generated images in complex datasets, making the synthesized results more reasonable for human observation, and thereby improves the perceptual quality. Table~\ref{ablation_fid} shows that the perceptual quality of synthesized results with WN is much better than models with BN and without BN, bringing $57.7\%$ and $43.6\%$ boosts to IS and FID respectively.

%------------------------
%-----------------------
%------------------------
\begin{figure*}[t!]
\begin{center}
\includegraphics[width=0.9\linewidth, angle=0]{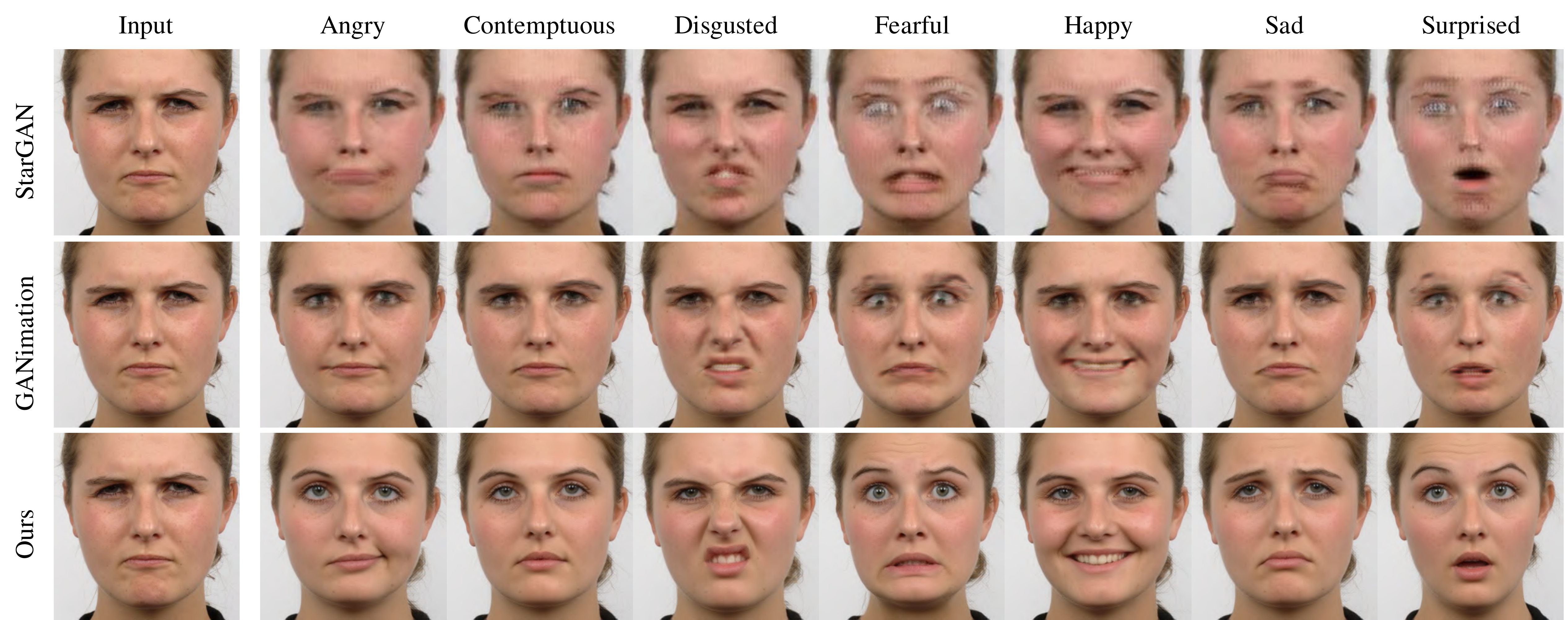}
\end{center}
\vspace{-0.5cm}
\caption{\label{comp_rafd}\small{Comparison with three state-of-the-art algorithms.(Zoom in for better details. Left three pictures are input faces, the right three lines are generated results of the three algorithms respectively. StarGAN~\cite{Choi_2018_CVPR} and GANimation~\cite{DBLP:conf/eccv/PumarolaAMSM18} are best current approaches under large poses, still show certain level of blur. Better zoom in to see the detail.}}
\vspace{-0.3cm}
\end{figure*}
%--------------------------------
\begin{figure}[h]
\begin{center}
\includegraphics[width=1.0\linewidth, angle=0]{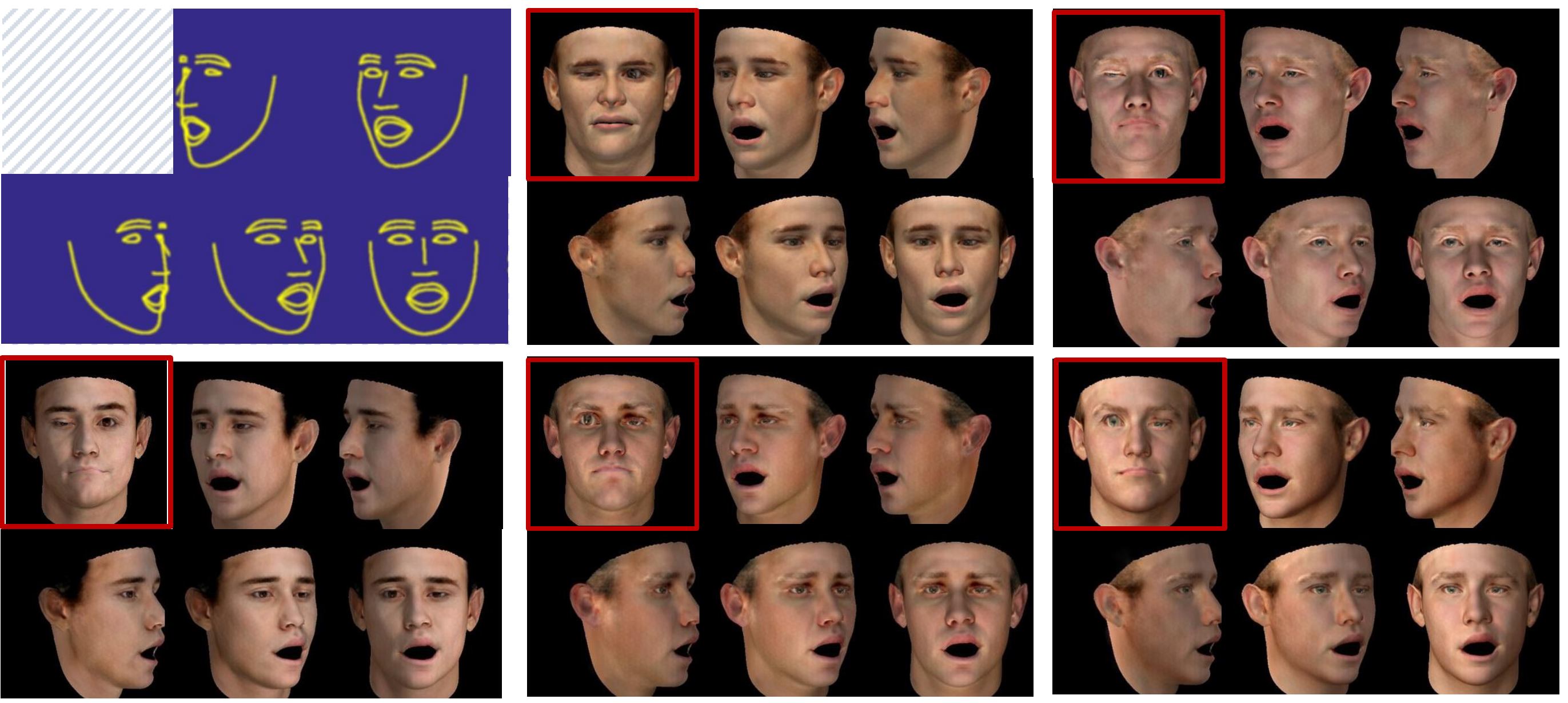}
\end{center}
\vspace{-0.5cm}
\caption{\label{sy_rotate}\small{Face rotation results on 3D synthetic face dataset.There are $6$ blocks, the upper left block represents $5$ target boundary maps. For each blocks, the upper left face with red box is the input image and the rest $5$ are synthesized results corresponding to $5$ boundary maps in the first block. Each source has different texture and lighting. Better zoom in to see the detail.}}
\vspace{-0.5cm}
\end{figure}
% Face rotation results on MultiPIE~\cite{gross2010multi} datasets

%----------------------------------
\section{Experiments}

Our framework provides a flexible way to manipulate an input face image into an arbitrary expression and pose under the control of boundary maps. In this section, we show qualitative and quantitative comparisons with state-of-the-art approaches in Sec ~\ref{SOTA}. Then, we perform self-evaluation to analyze the key components of our model in Sec~\ref{abl_sec}. Finally, we discuss the limitation of our approach in Sec~\ref{fail_sec}. All experiments are conducted using the model output from unseen images during the training phase.

\noindent
\textbf{Implementation Details.}
Before training, all images are aligned and cropped to $256\times256$ resolution. Facial landmarks of each image are obtained using an open-source pre-trained model. Then, landmarks are interpolated to get the facial boundary map. For the clustering, we choose $k=8$ as the number of clusters and K-means as the clustering algorithm to get clusters of boundaries. For more details, please refer to the appendix.

\noindent
\textbf{Datasets.} We mainly conduct experiments on RaFD~\cite{langner2010presentation},  MultiPIE~\cite{gross2010multi}, and CelebA~\cite{liu2015faceattributes} datasets that cover both in-door and in-the-wild setting. A 3D synthesized face dataset is also introduced to further evaluate the performance of the proposed method on facial texture detail,~\eg, illumination, complexion, and wrinkle. We use \textbf{hand} and \textbf{cat} as non-human datasets. The landmarks of hand and cat dataset are obtained through pre-trained hand detector and human annotation, respectively.

For each datasets, $90\%$ identities are used for training. and the left $10\%$ are fed into the model for testing. Additional quantitative and qualitative results on other datasets are shown in the appendix due to space limits.

\noindent
\textbf{Baselines.}
We compare our model with three state-of-the-art GAN-based algorithms: StarGAN~\cite{Choi_2018_CVPR}, GANimation~\cite{DBLP:conf/eccv/PumarolaAMSM18}, and pix2pixHD~\cite{Wang_2018_CVPR}. For a fair comparison, we
train these models using the implementations provided by the authors, and adopt publicly available pre-trained models to obtain conditional input, like Action Units (AUs) or landmarks for training and testing the models.

%--------------------------------
\begin{figure}[h]
\begin{center}
\includegraphics[width=1.0\linewidth, angle=0]{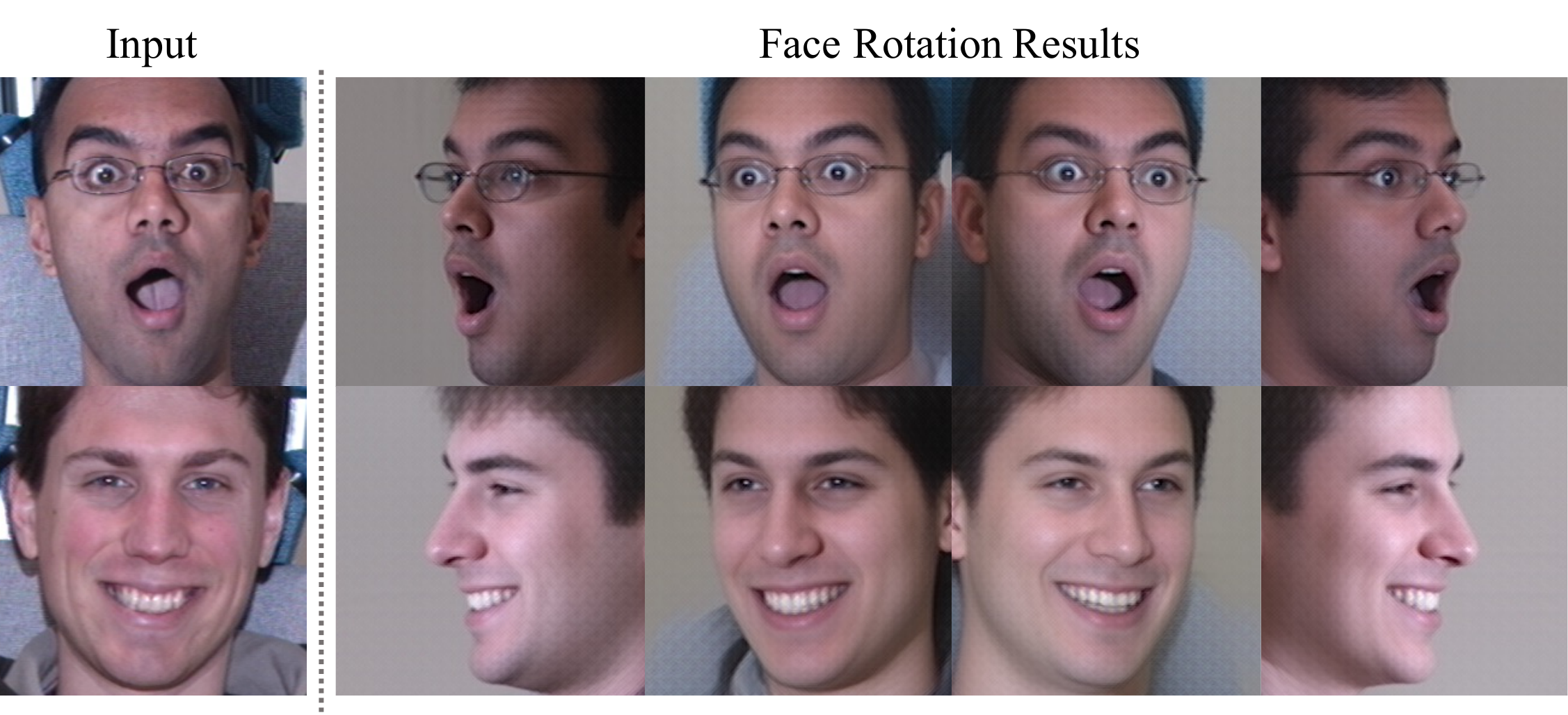}
\end{center}
\vspace{-0.6cm}
\caption{\label{rotate}\small{Face rotation results on MultiPIE~\cite{gross2010multi} dataset.}}
\vspace{-0.5cm}
\end{figure}
\noindent
\textbf{Performance metrics.}
For quantitative comparison, we evaluate three aspects, ~\ie the realism, the perceptual quality and the diversity, for the synthesis results. For the measurement of perceptual quality and diversity, we use \textbf{Fr\'echet Inception Distance} (FID, lower value indicates better quality)~\cite{heusel2017gans} and \textbf{Inception Score} (IS, higher value indicates better quality)~\cite{salimans2016improved} metrics. For realism, we use Amazon Mechanical Turk (AMT) to compare the perceived visual fidelity of our
method against existing approaches. We report \textbf{TS} (TrueSkill)~\cite{herbrich2007trueskill} and \textbf{FR} (Fool Rate, the estimated probability that generated images succeed in fooling the user) using data gathered from 25 participants per algorithm. Each participant is asked to complete 50 trials.

\subsection{\label{SOTA}Comparison with Existing Work}

\vspace{-0.2cm}
\subsubsection{Qualitative Comparisons}

\noindent
\textbf{Facial Expression Editing.} We conducted face manipulation in comparison with StarGAN, GANimation, and pix2pix on Rafd with 7 typical expressions. As shown in Fig.~\ref{comp_rafd}, previous leading methods are fragile when dealing with an exaggerated expression such as ~~``disgusted'' and ~~``fearful'' with $256\times256$ resolution. On the contrary, our method can get rid of blurry artifacts as well as maintaining facial details, owing to finely disentangled latent space and quality-refinement schema. It is worth noting that our results are far better than all of the GAN-based baselines especially in the detailed texture of mouth and eyes, which bring much higher perceptual quality. Quantitative evaluation in Sec.~\ref{quant} further demonstrates this observation.

\noindent
\textbf{Face Rotation.} We validate the ability of our model to face rotation task with an arbitrary oriented pose. Note that there is no requirement of any paired training samples and external supervision in our setting, which is different from the state-of-the-art methods such as~\cite{Huang_2017_ICCV,hu2018pose}. First, we conducted qualitative experiments on 3D synthetic face dataset to evaluate the detailed light/texture-preserving nature of our model. The driving landmark maps of different directions are obtained by manipulating the 3D landmark map to new expression and re-projecting it into different 2D views. As shown in Fig.~\ref{sy_rotate}, factors including complexion, texture, and lighting can be well preserved. This phenomenon testify the effectiveness of disentangling mechanism on our model. Next, we evaluate the effectiveness on real dataset. As shown in Fig.~\ref{rotate}, even under $90^\circ$, our model could still generate high fidelity and photo-realistic results in a simple weakly-supervised fashion.
\noindent
\begin{figure}[h]
\begin{center}
\includegraphics[width=0.9\linewidth, angle=0]{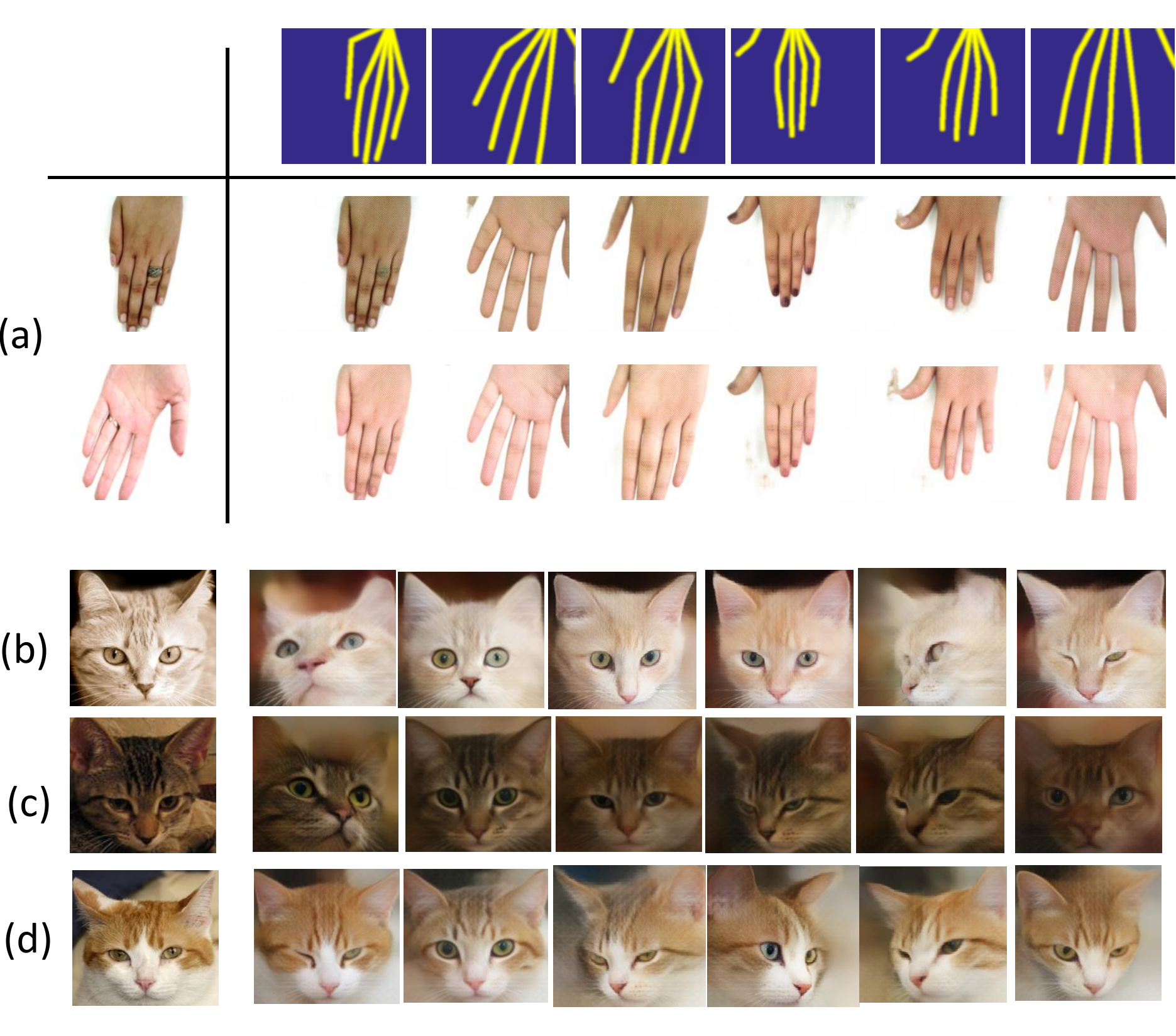}
\end{center}
\vspace{-0.6cm}
\caption{\label{cat}\small{Experiments on Hands and Cats datasets. (a) shows hands manipulation results. On the left are source hands. On the right, the first row presents target hand skeleton and the next two rows represent generated samples respectively. (b)(c)(d) are manipulated results of three cats. Each row presents a source cat image and $6$ manipulated results.}}
\vspace{-0.5cm}
\end{figure}

\subsubsection{\label{quant}Quantitative Comparisons}
First, we evaluate the perceptual quality and diversity of the generated images. As shown in Table~\ref{fid_sota}, our method outperforms the current state-of-the-art methods by a large margin on both measurements. Our FID and IS is 1.3 and 1.1 times better than the previous leading method, respectively.

Then, we use Amazon Mechanical Turk (AMT) to compare the perceived visual reality of our method against existing approaches. Table~\ref{user_study} reports results of AMT perceptual realism task. We find that our method can fool participants significantly better than other methods. Also, regarding TrueSkill, our model is more likely to gain users' preference, which further supports that our method surpasses those baselines by a large gap in terms of generated reality.

%-------------------
\begin{table}[]%\[!htb\]
\begin{center}

      \resizebox{0.6\columnwidth}{!}{
		\begin{tabular}{c|c|c}
		\Xhline{1.2pt}
		Model  & FID & IS \\
		\Xhline{1.2pt}
		Real Data  & 0.000 & 1.383 \\
		\hline
		pix2pixHD~\cite{Wang_2018_CVPR}  & 75.376 & 0.875 \\
		StarGAN~\cite{Choi_2018_CVPR}  & 56.937 & 1.036 \\
		GANimation\cite{DBLP:conf/eccv/PumarolaAMSM18}  & 34.360 & 1.112 \\
		\hline
		\textbf{Ours} & \textbf{25.069} & \textbf{1.237} \\
		\Xhline{1.2pt}
		\end{tabular}
      }
      \vspace{-0.2cm}
      \caption{\label{fid_sota}\small{Quantitative comparison with state-of-the-art on RaFD dataset with FID and IS metrics.}}
      \vspace{-0.8cm}
      \end{center}
\end{table}
%---------------
\begin{table}[h]
\begin{center}\begin{tabular}{c|c|c}
\Xhline{1.2pt}
Model  & Fool  Rate (\%) & TrueSkill \\
\hline
StarGAN~\cite{Choi_2018_CVPR}  & $3\% \pm 0.4\%$ & $18.1 \pm 0.9$ \\
pix2pixHD~\cite{Wang_2018_CVPR} & $4.8\% \pm 0.9\%$ & N/A \\
GANimation~\cite{DBLP:conf/eccv/PumarolaAMSM18}  & $7.0\% \pm 1.2\%$ & $24.4 \pm 0.8$ \\
\hline
\textbf{Ours}  & $\textbf{36.4\%} \pm 2.8\%$ & $\textbf{32.6} \pm 0.9$ \\
\Xhline{1.2pt}
\end{tabular}
\end{center}
\vspace{-0.6cm}
\caption{\label{user_study}\small{Evaluation of User Study, compare with state-of-the-arts}}
\vspace{-0.3cm}
\end{table}
%------------------

  \begin{table}[h]{}
  \begin{center}

        \resizebox{0.6\columnwidth}{!}{
		\begin{tabular}{c|c|c}
		\Xhline{1.2pt}
		Model  & FID & IS \\
		\Xhline{1.2pt}
		Real Data  & 0.000 & 2.662 \\
		\hline
		Ours w/o KL  & 56.275 & 1.863 \\
		Ours w/o GMM  & 62.657 & 1.792 \\
		Ours w/o PS  & 71.309 & 1.065 \\
		Ours w/o WN & 65.309 & 1.365 \\
		\hline
		\textbf{Ours} & \textbf{36.820} & \textbf{2.152} \\
		\Xhline{1.2pt}
		\end{tabular}
      }
    \vspace {-0.3cm}
    \caption{\label{ablation_fid}\small{} Ablation study on CelebA. }
    \vspace {-0.2cm}
      \end{center}
          \vspace {-0.55cm}
\end{table}

%----------------------
\begin{figure}[t!]
\begin{center}
\includegraphics[width=1\linewidth, angle=0]{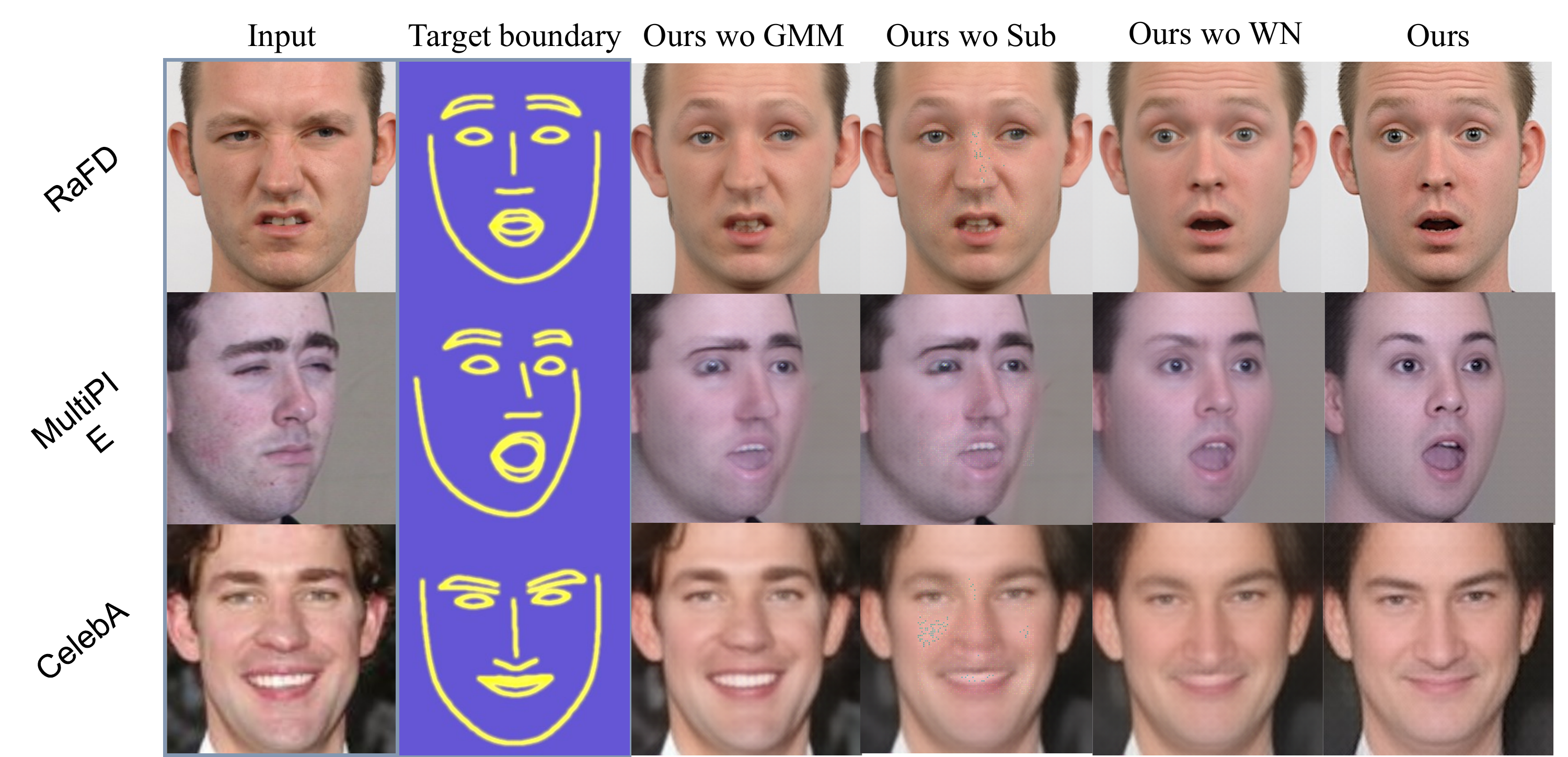}
\end{center}
\vspace{-0.6cm}
\caption{\label{ablation}\small{Ablation study on RafD, MultiPIE and CelebA datasets.}}
\vspace{-0.2cm}
\end{figure}
%-------------------

\subsection{\label{abl_sec}Ablation Study}
\indent
We conduct ablation study to analyze the contribution of individual components in the proposed method.

\noindent
\textbf{Qualitative ablation.} As shown in Fig.~\ref{ablation}, without GMM and KL, target with large spatial movements such as opening the mouth and turning faces around could not be generated. Without pixel shuffle, there are artifacts on generated faces. Without WN, the visual quality will be constrained.

\noindent
\textbf{Quantitative ablation.} We evaluate each variant based on the quality of their generated samples with FID and IS metrics. As shown in Table~\ref{ablation_fid}, the introduced GMM, pixel shuffle and weight normalization bring great improvement in terms of FID and IS score. Inferred from the above observations, each component has a different role in our method. Removing any of them leads to a performance drop.

\begin{figure}[t!]
\begin{center}
\includegraphics[width=1\linewidth, angle=0]{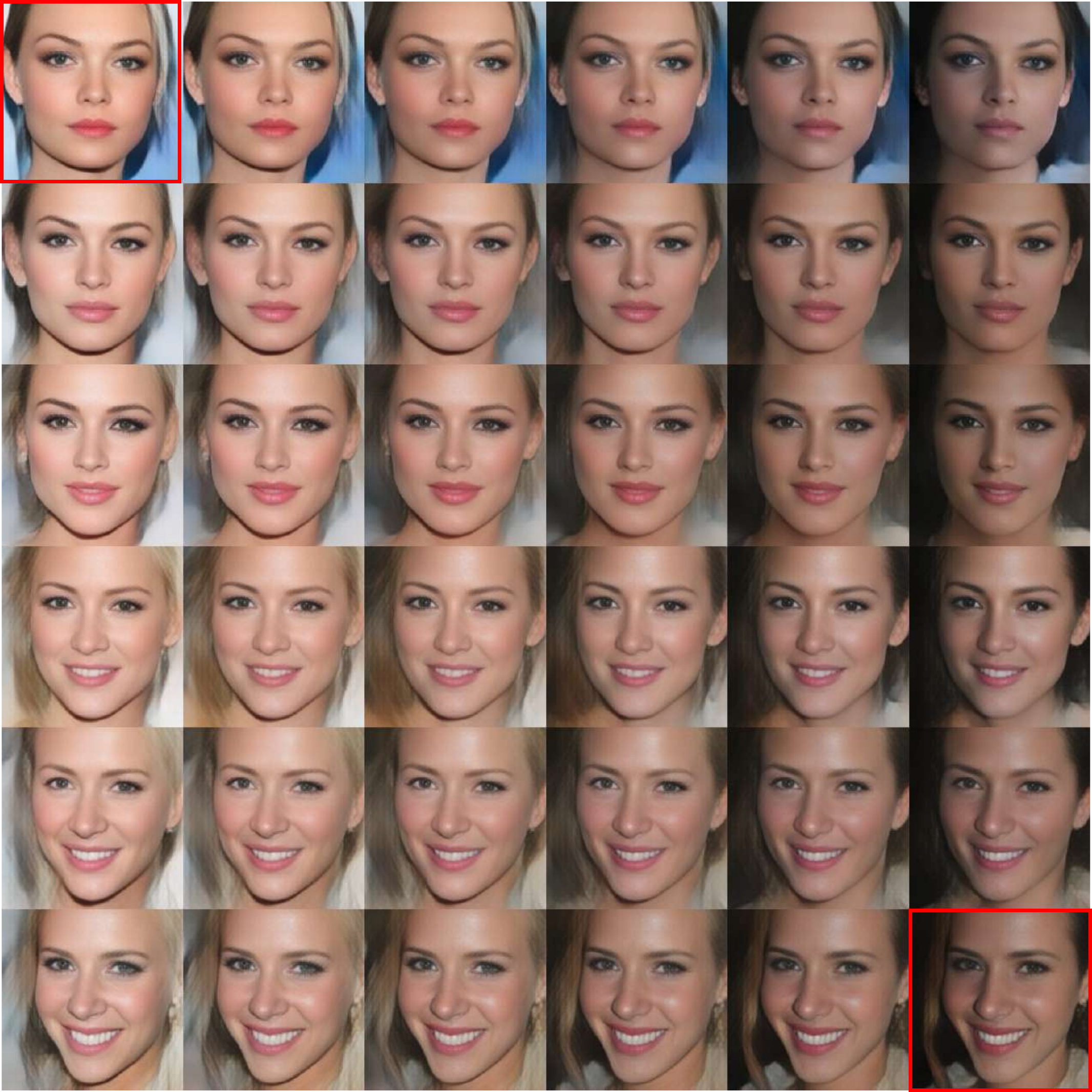}
\end{center}
\vspace{-0.5cm}
\caption{\label{interpolation}\small{Interpolation results on CelebA. Upper left and lower right are two real images, each row and line represents linear Interpolation on latent appearance and structure vector, respectively.}}
\vspace{-0.4cm}
\end{figure}

\noindent
\textbf{Interpolation Results.}
In order to validate that the feature distribution our model learned is dense and distinct, both the appearance and structure in the generated images should change continuously with the latent vector respectively. Our model demonstrates photo-realistic results in Fig.\ref{interpolation} via interpolation between disparate samples, suggesting that it has great generalization and robustness, instead of simply memorizing training data.

\begin{figure}[h]
\begin{center}
\includegraphics[width=1.0\linewidth, angle=0]{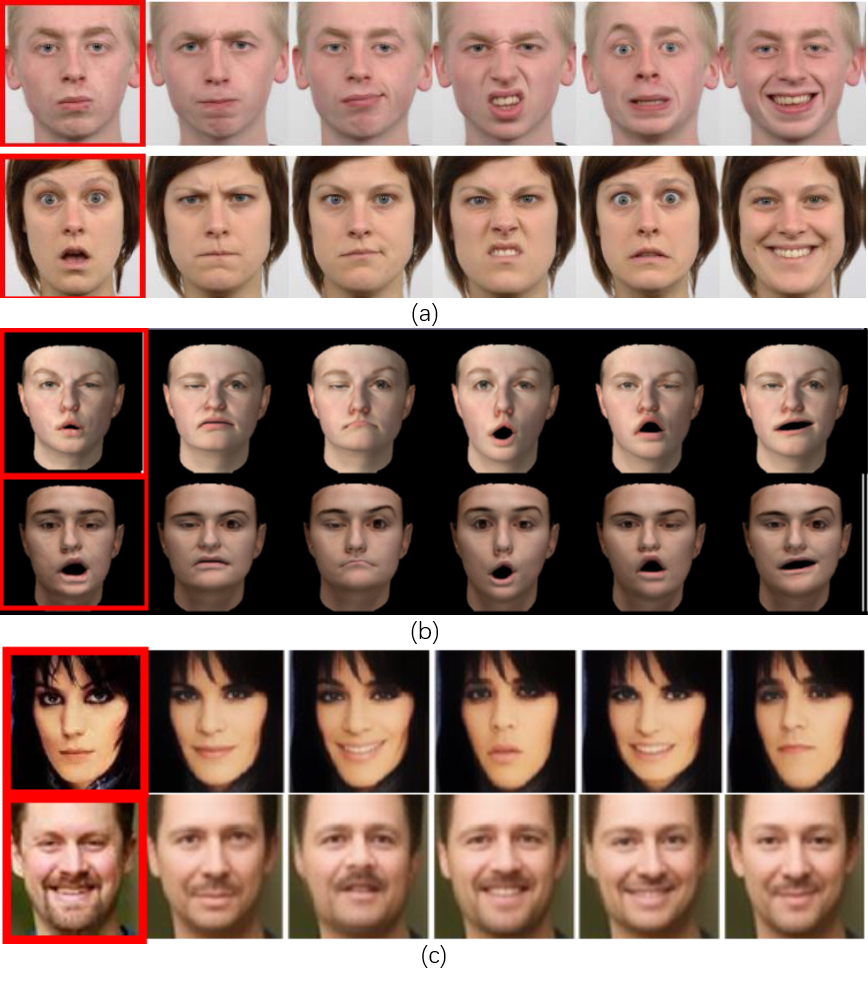}
\end{center}
\vspace{-0.6cm}
\caption{\label{identity}\small{Identity preserving evaluation on three experiment settings.}}
\vspace{-0.6cm}
\end{figure}

\noindent
\textbf{Identity Preserving Problem.}
Identity preserving is a long standing challenge on face generation domain. Previous leading methods~\cite{Bao_2018_CVPR,Shen_2018_CVPR} usually address this problem by extending the network with an identity classifier to constrain the variety of synthesized face. To study the influence of different factors to facial identity on our framework, we conduct three experiment settings on RafD, 3D synthetic face and EmotionNet datasets, as shown in Fig.~\ref{identity}. We find that when modifying the landmarks from source image (Fig.~\ref{identity}(a)), or using a boundary with similar facial contour from other person (Fig.~\ref{identity}(b)) as the conditional boundary map, the model could synthesize faces with both well preserved identity and high fidelity. Conversely, when the facial contour of conditional boundary map is significantly different with source image(Fig.~\ref{identity}(c)), the identity of synthesized result tend to be diverse.  These observation also indicates that facial identity information is mainly encoded in structure space, as demonstrated in~\cite{DBLP:journals/tvcg/CaoWZTZ14,Deng_2018_CVPR}.

\subsection{\label{fail_sec}Limitations and Failure Cases}

\begin{figure}[h]
\begin{center}
\includegraphics[width=\linewidth, angle=0]{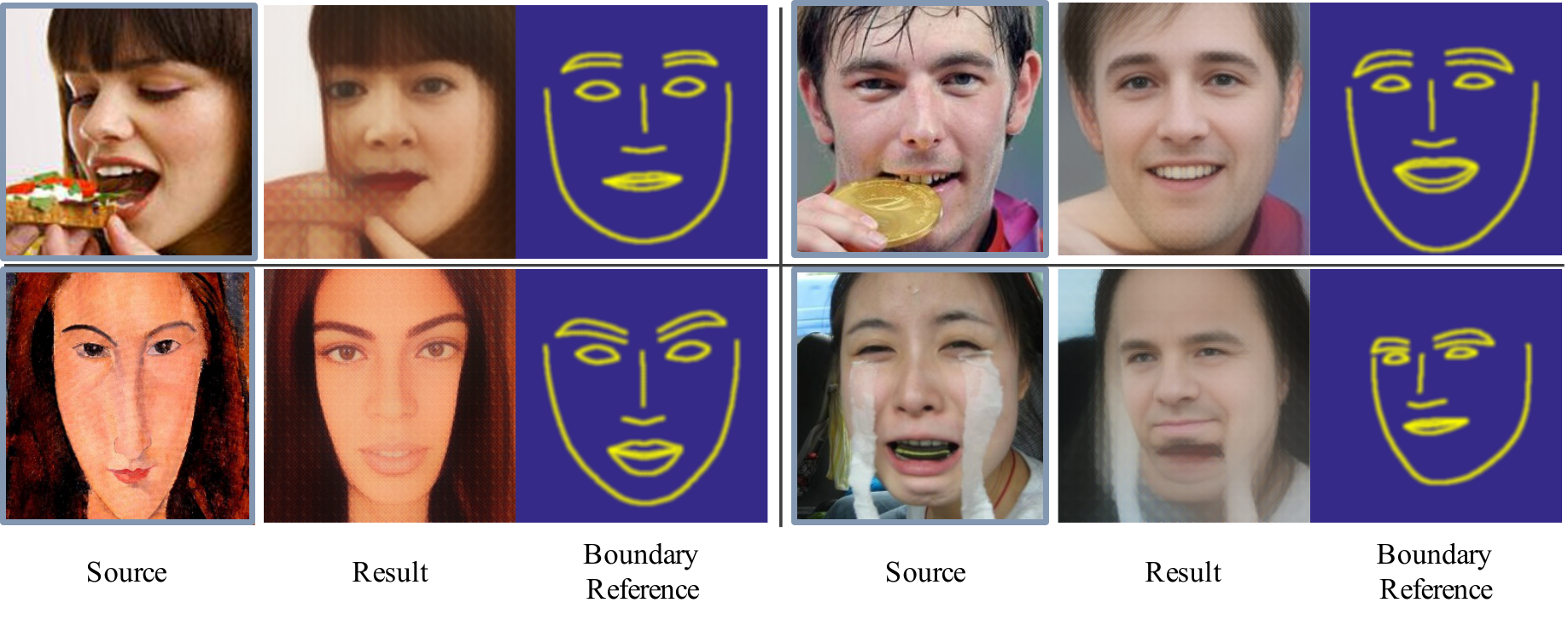}
\end{center}
\vspace{-0.6cm}
\caption{\label{failurcases}\small{Failure Cases. All four failure cases are selected from CelebA and EmotionNet dataset. We represent the source image at left, followed by manipulation result and its boundary reference.}}
\vspace{-0.5cm}
\end{figure}

We show four categories of failure cases in     Fig.~\ref{failurcases}, all of them are representative cases that challenge the limit of our model. Specifically, the first case (the top left one) is related to rare data. As cakes are seldom seen in our training samples, the model may not be able to maintain sample's semantics and tend to be blurry. A similar problem occurs in the top right picture where some source parts in source image are occluded. Our model tends to be confused when source image are occluded and may simply move the occlusions from the source due to strong structure coherence in model design. Another challenging case that our model couldn't handle well is special style. As shown in Fig.~\ref{failurcases} lower left, characteristics from the source style are likely to be lost when referred to a boundary from human face. As the estimated landmarks may strongly correlate to some attributes such as gender and head pose, simply imposed with such a given boundary, the result image are likely to simply morph those attributes with our target appearance as shown in lower right part in Fig.~\ref{failurcases}. By simply morphing the structure, the target result are unnatural and losing identity.

% \subsection{Application-Data augmentation for Face Alignment}

% One application our approach could perform is using generated samples to aid face alignment training. We use faces and their annotations from 300W~~\cite{sagonas2013300} for training. Then randomly sample a face and unpaired landmarks from the dataset, generating the face under that sampled pose and using input landmarks as its label. Training Details can be found in the Appendix. Table~\ref{tb_300w_fullset} shows the results of face alignment with data augmentations using generated images. By utilizing generated samples, our model surprisingly reach state-of-the-art result on 300W 68 points benchmark without using extra human-annotated data.

\noindent
\section{Conclusion}
In this paper, we propose an additive focal variational auto-encoder (AF-VAE) framework for face manipulation which is capable of modeling the complex interaction between facial structure and appearance. A light-weight architecture designed on HVS basis empowers better synthetic results. It advances current works in face synthesis on both generation quality and adaptation to extreme manipulation settings, as well as its simple structure and stable training procedure. We hope our work could illuminate more trails in this direction.
{\small
\bibliographystyle{ieee_fullname}
\bibliography{egbib}

\begin{thebibliography}{10}\itemsep=-1pt

\bibitem{DBLP:conf/iccv/BaoCWLH17}
Jianmin Bao, Dong Chen, Fang Wen, Houqiang Li, and Gang Hua.
\newblock {CVAE-GAN:} fine-grained image generation through asymmetric
  training.
\newblock In {\em ICCV}, 2017.

\bibitem{Bao_2018_CVPR}
Jianmin Bao, Dong Chen, Fang Wen, Houqiang Li, and Gang Hua.
\newblock Towards open-set identity preserving face synthesis.
\newblock In {\em CVPR}, 2018.

\bibitem{blanz1999morphable}
Volker Blanz and Thomas Vetter.
\newblock A morphable model for the synthesis of 3d faces.
\newblock In {\em Proceedings of the 26th annual conference on Computer
  graphics and interactive techniques}, 1999.

\bibitem{brock2018large}
Andrew Brock, Jeff Donahue, and Karen Simonyan.
\newblock Large scale gan training for high fidelity natural image synthesis.
\newblock {\em arXiv preprint arXiv:1809.11096}, 2018.

\bibitem{DBLP:journals/tvcg/CaoWZTZ14}
Chen Cao, Yanlin Weng, Shun Zhou, Yiying Tong, and Kun Zhou.
\newblock Facewarehouse: {A} 3d facial expression database for visual
  computing.
\newblock {\em {IEEE} Trans. Vis. Comput. Graph.}, 20(3):413--425, 2014.

\bibitem{chen2017photographic}
Qifeng Chen and Vladlen Koltun.
\newblock Photographic image synthesis with cascaded refinement networks.
\newblock In {\em ICCV}, 2017.

\bibitem{chen2018facelet}
Ying-Cong Chen, Huaijia Lin, Michelle Shu, Ruiyu Li, Xin Tao, Xiaoyong Shen,
  Yangang Ye, and Jiaya Jia.
\newblock Facelet-bank for fast portrait manipulation.
\newblock In {\em CVPR}, 2018.

\bibitem{DBLP:journals/corr/abs-1801-07632}
Zeyuan Chen, Shaoliang Nie, Tianfu Wu, and Christopher~G. Healey.
\newblock High resolution face completion with multiple controllable attributes
  via fully end-to-end progressive generative adversarial networks.
\newblock {\em arXiv preprint arXiv:1801.07632}, 2018.

\bibitem{Choi_2018_CVPR}
Yunjey Choi, Minje Choi, Munyoung Kim, Jung-Woo Ha, Sunghun Kim, and Jaegul
  Choo.
\newblock Stargan: Unified generative adversarial networks for multi-domain
  image-to-image translation.
\newblock In {\em CVPR}, 2018.

\bibitem{Deng_2018_CVPR}
Jiankang Deng, Shiyang Cheng, Niannan Xue, Yuxiang Zhou, and Stefanos
  Zafeiriou.
\newblock Uv-gan: Adversarial facial uv map completion for pose-invariant face
  recognition.
\newblock In {\em CVPR}, 2018.

\bibitem{esser2018variational}
Patrick Esser, Ekaterina Sutter, and Bj{\"o}rn Ommer.
\newblock A variational u-net for conditional appearance and shape generation.
\newblock In {\em CVPR}, 2018.

\bibitem{gross2010multi}
Ralph Gross, Iain Matthews, Jeffrey Cohn, Takeo Kanade, and Simon Baker.
\newblock Multi-pie.
\newblock {\em Image and Vision Computing}, 2010.

\bibitem{gulrajani2017improved}
Ishaan Gulrajani, Faruk Ahmed, Martin Arjovsky, Vincent Dumoulin, and Aaron~C
  Courville.
\newblock Improved training of wasserstein gans.
\newblock In {\em NIPS}, 2017.

\bibitem{herbrich2007trueskill}
Ralf Herbrich, Tom Minka, and Thore Graepel.
\newblock Trueskill™: a bayesian skill rating system.
\newblock In {\em NIPS}, 2007.

\bibitem{heusel2017gans}
Martin Heusel, Hubert Ramsauer, Thomas Unterthiner, Bernhard Nessler, and Sepp
  Hochreiter.
\newblock Gans trained by a two time-scale update rule converge to a local nash
  equilibrium.
\newblock In {\em NIPS}, 2017.

\bibitem{hu2018pose}
Yibo Hu, Xiang Wu, Bing Yu, Ran He, and Zhenan Sun.
\newblock Pose-guided photorealistic face rotation.
\newblock In {\em CVPR}, 2018.

\bibitem{huang2018introvae}
Huaibo Huang, Zhihang Li, Ran He, Zhenan Sun, and Tieniu Tan.
\newblock Introvae: Introspective variational autoencoders for photographic
  image synthesis.
\newblock {\em arXiv preprint arXiv:1807.06358}, 2018.

\bibitem{Huang_2017_ICCV}
Rui Huang, Shu Zhang, Tianyu Li, and Ran He.
\newblock Beyond face rotation: Global and local perception gan for
  photorealistic and identity preserving frontal view synthesis.
\newblock In {\em ICCV}, 2017.

\bibitem{ioffe2015batch}
Sergey Ioffe and Christian Szegedy.
\newblock Batch normalization: Accelerating deep network training by reducing
  internal covariate shift.
\newblock {\em arXiv preprint arXiv:1502.03167}, 2015.

\bibitem{DBLP:conf/cvpr/IsolaZZE17}
Phillip Isola, Jun{-}Yan Zhu, Tinghui Zhou, and Alexei~A. Efros.
\newblock Image-to-image translation with conditional adversarial networks.
\newblock In {\em CVPR}, 2017.

\bibitem{Kingma2014}
Diederik~P. Kingma and Max Welling.
\newblock Auto-encoding variational bayes.
\newblock In {\em ICLR}, 2014.

\bibitem{Kossaifi_2018_CVPR}
Jean Kossaifi, Linh Tran, Yannis Panagakis, and Maja Pantic.
\newblock Gagan: Geometry-aware generative adversarial networks.
\newblock In {\em CVPR}, 2018.

\bibitem{langner2010presentation}
Oliver Langner, Ron Dotsch, Gijsbert Bijlstra, Daniel~HJ Wigboldus, Skyler~T
  Hawk, and AD Van~Knippenberg.
\newblock Presentation and validation of the radboud faces database.
\newblock {\em Cognition and emotion}, 2010.

\bibitem{Lin_2018_CVPR}
Kwan-Yee Lin and Guanxiang Wang.
\newblock Hallucinated-iqa: No-reference image quality assessment via
  adversarial learning.
\newblock In {\em CVPR}, 2018.

\bibitem{Liu_2017_ICCV}
Xialei Liu, Joost van~de Weijer, and Andrew~D. Bagdanov.
\newblock Rankiqa: Learning from rankings for no-reference image quality
  assessment.
\newblock In {\em ICCV}, 2017.

\bibitem{liu2015faceattributes}
Ziwei Liu, Ping Luo, Xiaogang Wang, and Xiaoou Tang.
\newblock Deep learning face attributes in the wild.
\newblock In {\em ICCV}, 2015.

\bibitem{lu2018attribute}
Yongyi Lu, Yu-Wing Tai, and Chi-Keung Tang.
\newblock Attribute-guided face generation using conditional cyclegan.
\newblock In {\em ECCV}, 2018.

\bibitem{ma2017pose}
Liqian Ma, Xu Jia, Qianru Sun, Bernt Schiele, Tinne Tuytelaars, and Luc
  Van~Gool.
\newblock Pose guided person image generation.
\newblock In {\em NIPS}, 2017.

\bibitem{ma2018disentangled}
Liqian Ma, Qianru Sun, Stamatios Georgoulis, Luc Van~Gool, Bernt Schiele, and
  Mario Fritz.
\newblock Disentangled person image generation.
\newblock In {\em CVPR}, 2018.

\bibitem{mao2017least}
Xudong Mao, Qing Li, Haoran Xie, Raymond~YK Lau, Zhen Wang, and Stephen~Paul
  Smolley.
\newblock Least squares generative adversarial networks.
\newblock In {\em ICCV}, 2017.

\bibitem{Pan_2018_CVPR}
Da Pan, Ping Shi, Ming Hou, Zefeng Ying, Sizhe Fu, and Yuan Zhang.
\newblock Blind predicting similar quality map for image quality assessment.
\newblock In {\em CVPR}, 2018.

\bibitem{DBLP:conf/eccv/PumarolaAMSM18}
Albert Pumarola, Antonio Agudo, Aleix~M. Martinez, Alberto Sanfeliu, and
  Francesc Moreno{-}Noguer.
\newblock Ganimation: Anatomically-aware facial animation from a single image.
\newblock In {\em ECCV}, 2018.

\bibitem{qianextending}
Shengju Qian, Wayne Wu, Yangxiaokang Liu, Beier Zhu, and Fumin Shen.
\newblock Extending the capacity of cvae for face synthesis and modeling.
\newblock In {\em NeurIPS Workshops}, 2018.

\bibitem{qiao2018geometry}
Fengchun Qiao, Naiming Yao, Zirui Jiao, Zhihao Li, Hui Chen, and Hongan Wang.
\newblock Geometry-contrastive generative adversarial network for facial
  expression synthesis.
\newblock {\em arXiv preprint arXiv:1802.01822}, 2018.

\bibitem{salimans2016improved}
Tim Salimans, Ian Goodfellow, Wojciech Zaremba, Vicki Cheung, Alec Radford, and
  Xi Chen.
\newblock Improved techniques for training gans.
\newblock In {\em NIPS}, 2016.

\bibitem{salimans2016weight}
Tim Salimans and Diederik~P Kingma.
\newblock Weight normalization: A simple reparameterization to accelerate
  training of deep neural networks.
\newblock In {\em NIPS}, 2016.

\bibitem{Shen_2018_CVPR}
Yujun Shen, Ping Luo, Junjie Yan, Xiaogang Wang, and Xiaoou Tang.
\newblock Faceid-gan: Learning a symmetry three-player gan for
  identity-preserving face synthesis.
\newblock In {\em CVPR}, 2018.

\bibitem{shi2016real}
Wenzhe Shi, Jose Caballero, Ferenc Husz{\'a}r, Johannes Totz, Andrew~P Aitken,
  Rob Bishop, Daniel Rueckert, and Zehan Wang.
\newblock Real-time single image and video super-resolution using an efficient
  sub-pixel convolutional neural network.
\newblock In {\em CVPR}, 2016.

\bibitem{DBLP:conf/eccv/ShuSGSPK18}
Zhixin Shu, Mihir Sahasrabudhe, Riza~Alp G{\"{u}}ler, Dimitris Samaras, Nikos
  Paragios, and Iasonas Kokkinos.
\newblock Deforming autoencoders: Unsupervised disentangling of shape and
  appearance.
\newblock In {\em ECCV}, 2018.

\bibitem{DBLP:conf/cvpr/ShuYHSSS17}
Zhixin Shu, Ersin Yumer, Sunil Hadap, Kalyan Sunkavalli, Eli Shechtman, and
  Dimitris Samaras.
\newblock Neural face editing with intrinsic image disentangling.
\newblock In {\em CVPR}, 2017.

\bibitem{sohn2015learning}
Kihyuk Sohn, Honglak Lee, and Xinchen Yan.
\newblock Learning structured output representation using deep conditional
  generative models.
\newblock In {\em NIPS}, 2015.

\bibitem{thies2016face2face}
Justus Thies, Michael Zollhofer, Marc Stamminger, Christian Theobalt, and
  Matthias Nie{\ss}ner.
\newblock Face2face: Real-time face capture and reenactment of rgb videos.
\newblock In {\em CVPR}, 2016.

\bibitem{nonlinear-3d-face-morphable-model}
Luan Tran and Xiaoming Liu.
\newblock Nonlinear 3d face morphable model.
\newblock In {\em CVPR}, 2018.

\bibitem{on-learning-3d-face-morphable-model-from-in-the-wild-images}
Luan Tran and Xiaoming Liu.
\newblock On learning 3d face morphable model from in-the-wild images.
\newblock {\em arXiv preprint arXiv:1808.09560}, 2018.

\bibitem{wang2017diverse}
Liwei Wang, Alexander Schwing, and Svetlana Lazebnik.
\newblock Diverse and accurate image description using a variational
  auto-encoder with an additive gaussian encoding space.
\newblock In {\em NIPS}, 2017.

\bibitem{wang2018vid2vid}
Ting-Chun Wang, Ming-Yu Liu, Jun-Yan Zhu, Guilin Liu, Andrew Tao, Jan Kautz,
  and Bryan Catanzaro.
\newblock Video-to-video synthesis.
\newblock In {\em NIPS}, 2018.

\bibitem{Wang_2018_CVPR}
Ting-Chun Wang, Ming-Yu Liu, Jun-Yan Zhu, Andrew Tao, Jan Kautz, and Bryan
  Catanzaro.
\newblock High-resolution image synthesis and semantic manipulation with
  conditional gans.
\newblock In {\em CVPR}, 2018.

\bibitem{wang2018every}
Wei Wang, Xavier Alameda-Pineda, Dan Xu, Pascal Fua, Elisa Ricci, and Nicu
  Sebe.
\newblock Every smile is unique: Landmark-guided diverse smile generation.
\newblock In {\em CVPR}, 2018.

\bibitem{Wiles18a}
Olivia Wiles, A Koepke, and Andrew Zisserman.
\newblock Self-supervised learning of a facial attribute embedding from video.
\newblock In {\em BMVC}, 2018.

\bibitem{wiles2018x2face}
Olivia Wiles, A~Sophia Koepke, and Andrew Zisserman.
\newblock X2face: A network for controlling face generation using images,
  audio, and pose codes.
\newblock In {\em ECCV}, 2018.

\bibitem{wayne2019transgaga}
Wayne Wu, Kaidi Cao, Cheng Li, Chen Qian, and Chen~Change Loy.
\newblock Transgaga: Geometry-aware unsupervised image-to-image translation.
\newblock In {\em CVPR}, 2019.

\bibitem{Wu_2018_CVPR}
Wayne Wu, Chen Qian, Shuo Yang, Quan Wang, Yici Cai, and Qiang Zhou.
\newblock Look at boundary: A boundary-aware face alignment algorithm.
\newblock In {\em CVPR}, 2018.

\bibitem{wayne2018reenactgan}
Wayne Wu, Yunxuan Zhang, Cheng Li, Chen Qian, and Chen~Change Loy.
\newblock Reenactgan: Learning to reenact faces via boundary transfer.
\newblock In {\em ECCV}, 2018.

\bibitem{yang2018pose}
Ceyuan Yang, Zhe Wang, Xinge Zhu, Chen Huang, Jianping Shi, and Dahua Lin.
\newblock Pose guided human video generation.
\newblock In {\em ECCV}, 2018.

\bibitem{towards-large-pose-face-frontalization-in-the-wild}
Xi Yin, Xiang Yu, Kihyuk Sohn, Xiaoming Liu, and Manmohan Chandraker.
\newblock Towards large-pose face frontalization in the wild.
\newblock In {\em ICCV}, 2017.

\bibitem{zhang2018perceptual}
Richard Zhang, Phillip Isola, Alexei~A Efros, Eli Shechtman, and Oliver Wang.
\newblock The unreasonable effectiveness of deep features as a perceptual
  metric.
\newblock In {\em CVPR}, 2018.

\end{thebibliography}
}

\end{document}

%%%%%%%%% TITLE
\title{Supplementary Material\\
Make a Face: Towards Arbitrary High Fidelity Face Manipulation}

\maketitle
\thispagestyle{empty}
%%%%%%%%% ABSTRACT

\begin{abstract}
This documents provides additional information regarding our main paper, network architecture and  detailed implementations . Furthermore, we provide extend results to illustrate the advantages of our method due to space limits.

\noindent
The supplementary material is organized as follows:
\begin{itemize}
\item Sec.~\ref{CVAE} illustrates the basic formulation and usage of C-VAE on solving our problems.
\item Sec.~\ref{Weight} presents effects of weight normalization in detail, including loss curves and further analysis.
\item Sec.~\ref{network} describes the network architecture used in experiments. 
\item Sec.~\ref{exp_setting} provides more experiments details, including detailed datasets descriptions, AMT perceptual scores used and specific implementations. 
\item Sec.~\ref{add_exp} shows additional experiments results on more datasets, including more general non-human cases. We also demonstrate more applications using our model, including face animation and video generation.
\end{itemize}
\end{abstract}

%%%%%%%%% BODY TEXT
\section{\label{CVAE}Modeling faces based on latent shape and appearance}

Given samples $x$ from a face dataset, it is straightforward to use VAEs which aim at modeling the data likelihood $p(x)$.
As VAEs assume that the data points $x$ around a low-dimensional manifold could be parameterized by latent embeddings $z$. The decoder is employed to obtain $x$ based on latent representation $z$. As a result, in order to make posterior $p(z|x)$ tractably computable, a encoder is applied to approximate it with a distribution $q(z|x)$. Thus, VAEs are based on the formulation:
\begin{equation}\label{vae}
\begin{split}
\log p(x) - D_{KL}[q(z|x),p(z|x)] & = \mathbb{E}_{q(z|x)}[\log p(x|z)] \\ & - D_{KL}[q(z|x),p(z)]
\end{split}
\end{equation}
However, for tasks like face manipulation or conditional image synthesis, synthesized images are conditioned by two factors $y$, $z$. Let $y$ denotes head pose information, including motion, expression, yaw angles and other spatial information. $z$ should refer to internal appearance information. Apparently, according to the formulation in equation~\ref{vae}, $z$ couldn't be well separated and thus are not able to satisfy our goal in dealing with tasks like face manipulation. 

As discussed in the main paper, in order to disentangle the two factors, $y$ and $z$ and learn the distribution $p(x|y,z)$. As $y$ could provide additional spatial information(landmarks in our case), the latent variable $z$ could be inferred by maximizing their conditional log likelihood:
 \begin{align}\label{elbo}
 log p(x|y) & = log\int_{z} p(x, z|y) dz \ge \mathbb{E}_{q} \frac{p(x,z|y)}{q(z|x,y)} \\ & = \mathbb{E}_{q}log \frac{p(z|,y,z)p(z|y)}{q(z|x,y)}
 \end{align}
According to equation~\ref{elbo}, the evidence lower bound (ELBO) now depends on the conditional prior $p(z|y)$. Thus more semantic correlations between spatial and appearance information could be captured. 
\section{\label{Weight}Weight Normalization and its Effects}

As mentioned before, for model training, weight normalization~\cite{salimans2016weight} is utilized. As has been tailored in GAN-based approaches, different normalization tricks are used mostly in order to stabilize adversarial training steps. 

For VAE-based generative models, which are particularly noise-sensitive, potential noises brought by normalization tricks~\cite{ioffe2015batch} would reduce the diversity of generated samples significantly, while the absence of normalization tricks would affect model's convergence. Hence we tailor this strategy to both \textit{encoder} and \textit{decoder} of our model. As can be seen from Fig.~\ref{train_curve}, models trained with weight normalization have faster convergence and lower reconstruction loss than the one without WN over training iterations. We also compare using 'vanilla' weight normalization with using weight normalization and sub pixel(pixel shuffle) convolution as ablative study. The blue curve represents using WN and PS is covered by the red one. It shows that weight normalization help training converge and enhance model's generative power. These curves also supports that pixel shuffle and weight normalization helps synthesis respectively in a different way as we conveyed in the main paper.

Specifically, Batch normalization re-calibrates the mean and variance of intermediate features to solve the problem of internal covariate shift during deep nets training, its formulation doesn't fit generative tasks very well with two drawbacks: on the one hand, for high-fidelity/resolution image synthesis, small mini-batch sizes are always applied due to limited GPU memory. On the other hand, especially for VAE-based 'noise-sensitive' variants, with imposed distribution constrains and noise gained by normalization tricks statistics, VAE's performance would be constrained significantly. 

\begin{figure}[]
\begin{center}
\includegraphics[width=1\linewidth, angle=0]{./imgs/supp/curve.png}
\end{center}
\caption{\label{train_curve}Training Curve w/wo pixel-shuffle and WN.}
\vspace{-0.5cm}
\end{figure}

\noindent
\textbf{Formulation} Assume the output \(\mathbf{y}\) is with the form:
\begin{equation}
\mathbf{y} = \mathbf{w} \cdot \mathbf{x} + b, 
\end{equation}
where \(\mathbf{w}\) is a k-dimensional vector representing weight, \(b\) is the bias term, \(\mathbf{x}\) is a k-dimensional input features. WN re-parameterizes the weight using
\begin{equation}
\mathbf{w} = \frac{g}{||\mathbf{v}||} \mathbf{v}, 
\end{equation}
where v is a k-dimensional vector, g is a scalar, and \(||\mathbf{v}||\) denotes the Euclidean norm of \(\mathbf{v}\). With this formalization, we will have \(||\mathbf{w}|| = g\), independent of parameters \(\mathbf{v}\). 

In practice, normalization tricks could stabilize GAN's training and aid model's performance. While for VAE training, which optimize the KL divergene where the loss quantities could be large through training, weight normalization is applied to avoid potentialities of collapse. Intuitively, WN is just a normalization trick and doesn't effect given representation power. This phenomenon has also been observed in image SR tasks~\cite{yu2018wide} recently. Compared to BN, Weight Normalization addresses these drawbacks using BN, also eases the difficulty of  training VAEs.

\section{\label{network}Network Architecture}
In this section, we provide details regarding the network architectures in our networks. Our network contains two branches of encoder and one decoder, which is illustrated in Fig.~\ref{architecture}. For all the experiments which synthesize the size of $256\times256$ images, we use $6$ residual blocks for down-sample. Detailed visualization can be found at Fig.~\ref{architecture} and specific choices of parameters can be found at Tables.~\ref{encoder}~\ref{decoder}.

\begin{figure}[htb]
\begin{center}
\includegraphics[width=0.8\linewidth, angle=0]{./imgs/supp/overview.png}
\end{center}
\vspace{-0.6cm}
\caption{\label{architecture}\small{Our basic model architecture. Skip connection on the structure branch is incorporated. As mentioned, we down-sample $6$ times for a image during our most experiments. During training, we sample the appearance distribution. }}
\vspace{-0.5cm}
\end{figure}

\begin{table}[htb]
\small
\begin{center}\begin{tabular}{c|c|c|c} 
\Xhline{1.2pt}
Layer  & Kernel Size & Output Channel & Output Size \\
\hline
Input  & - & 3 & 256 \\
\hline
Convolution & 3 & 64 & 256 \\
Pooling  & 2 & 64 & 128 \\
\hline
Convolution & 3 & 128 & 128 \\
Pooling  & 2 & 128 & 64 \\
\hline
Convolution & 3 & 256 & 64 \\
Pooling  & 2 & 256 & 32 \\
\hline
Convolution & 3 & 512 & 32 \\
Pooling  & 2 & 512 & 16 \\
\hline
Convolution & 3 & 512 & 16 \\
Pooling  & 2 & 512 & 8 \\
\hline
Convolution & 3 & 512 & 8 \\
Pooling  & 2 & 512 & 4 \\
\hline
\textbf{Output}  & - & 512 & 4 \\
level1 output & - & 512 & 8 \\
\Xhline{1.2pt}
\end{tabular}
\end{center}
\vspace{-0.5cm}
\caption{\label{encoder}\small{Details parameters of encoders, upper-branch encoder share the same architecture of the lower one. }}
\vspace{-0.5cm}
\end{table}

\begin{table}[htb]
\small
\begin{center}\begin{tabular}{c|c|c|c} 
\Xhline{1.2pt}
Layer  & Kernel Size & Output Channel & Output Size \\
\hline
Input  & - & 512 & 4 \\
\hline
Convolution & 3 & 2048 & 4 \\
Pixel Shuffle  & - & 512 & 8 \\
\hline
Concat & - & 1024 & 8 \\
\hline
Convolution & 3 & 2048 & 8 \\
Pixel Shuffle  & - & 512 & 16 \\
\hline
Concat & - & 1024 & 16 \\
\hline
Convolution & 3 & 1024 & 16 \\
Pixel Shuffle  & - & 256 & 32 \\
\hline
Concat & - & 512 & 32 \\
\hline
Convolution & 3 & 512 & 32 \\
Pixel Shuffle  & - & 128 & 64 \\
\hline
Concat & - & 256 & 64 \\
\hline
Convolution & 3 & 256 & 64 \\
Pixel Shuffle  & - & 64 & 128 \\
\hline
Concat & - & 128 & 128 \\
\hline
Convolution & 3 & 12 & 128 \\
Pixel Shuffle  & - & 3 & 256 \\
\Xhline{1.2pt}
\end{tabular}
\end{center}
\vspace{-0.5cm}
\caption{\label{decoder}\small{Details of decoders. Note that for fused feature is not included in the parameters since it depends on how many levels of latent features to be fused.  }}
\vspace{-0.5cm}
\end{table}
\section{\label{exp_setting}Experiments Details}
In this section, more details of experiments are provided. We first introduce all datasets we used in details in Sec.~\ref{dataset}. Then specific settings and usage of evaluation protocols are shown in Sec.~\ref{evaluation}. We also provide more implementation details in Sec.~\ref{implementation}.

\subsection{\label{dataset}Dataset descriptions}
We evaluate our model on RafD~\cite{langner2010presentation}, MultiPIE~\cite{gross2010multi}, CelebA~\cite{liu2015faceattributes} and 3D synthesized datasets, containing both indoor and in-the-wild scenarios. We also conduct experiments on none-human datasets, including hands~\cite{afifi2017gender} and cats datasets, in order to demonstrate the generalizability of our approach. Followings are detailed descriptions of each dataset we used in experiment.

For real-word human faces datasets: (1)\textbf{RafD}~\cite{langner2010presentation} dataset consists of $4,824$ images collected from $67$ participants with $8$ facial expressions in three different gaze directions for each person. $(2)$ \textbf{MultiPIE}~\cite{gross2010multi} consists of $20$ illumination conditions, $13$ poses within $90$ yaw angles and $6$ expressions of 337 subjects. These two datasets are captured in controlled environments. $(3)$ \textbf{CelebA}~\cite{liu2015faceattributes} is a large-scale face attributes dataset in uncontrolled environment, which contains $202,599$ face images of celebrities with large pose variations and background clutter. 

For \textbf{3D synthetic} face data, we employ 3DMM~\cite{paysan20093d} to create $17920$ distinct faces image with 10 shapes indicate 10 different types of facial structures. Each face has been rendered at $8$ types of random texture and $4$ different lighting environment have been used. Also, $8$ expressions are chose for each faces. The orientation of faces is allowed to vary in azimuth from $0^{\circ}$ to $180^{\circ}$ by increments of $30^{\circ}$.  

For non-human datasets. We use \textbf{hand} and \textbf{cat}. (1) \textbf{Hands}. The landmarks of each hands is obtained using pre-trained hand detector. Then the skeleton is interpolated using the landmarks obtained. (2)\textbf{Cats}, we crawled cats image from the Internet. Then, each cat face is annotated with $18$ landmarks.

For each datasets, $90\%$ identities are used for training. and the left $10\%$ are fed into the model for testing. 

\subsection{\label{evaluation}Evaluation Metrics}

We evaluate the realism perceptual quality and diversity of synthesis results using different metrics. Human subjective study on Amazon Mechanical Turk (AMT) are conducted for realism measurement. Detailed illustrations of metrics we used are provided:

\textbf{TS} (TrueSkill)~\cite{herbrich2007trueskill} and \textbf{FR} (Fool Rate) are reported. Specifically, \textbf{TS} is a skill-based ranking system containing multiple players that allows us to compare our method with other methods. In practice, we randomly select two images generated by three algorithms. The participant is asked to decide which image demonstrates better quality. \textbf{FR} is used to evaluate the fidelity of generated images based on direct comparison with real images. Given two samples, a participant is asked to distinguish the real one. Note that since pix2pixHD~\cite{Wang_2018_CVPR} solely depends on condition and are not able to preserve the reference person. Thus we didn't conduct Fool Rate evaluation using pix2pixHD. 

\subsection{\label{implementation}Implementation Details}
Before training, all faces  are cropped and aligned to $256\times256$.  For the network structure, details can be found at Fig.~\ref{architecture}. PReLU~\cite{trottier2017parametric} is used as the activation function. Each BN module in residual block is replaced by WN module and sub-pixel convolution is used for up-sampling. 

At training, we used Adam~\cite{kingma:adam} optimizer with learning rate of 0.001, beta1 0.5, beta2 0.999 and batch size 4 on a single GTX TITANX GPU. For the feature matching loss using VGG-19, we simply follow perceptual loss~\cite{johnson2016perceptual}. The coefficient of KL divergence loss is set to $1$. 

\section{\label{add_exp}Additional Experiments Results}

Followings are additional experiments for comprehensive demonstration. 

\iffalse
\begin{figure*}[h]
\begin{center}
\includegraphics[width=1.0\linewidth, angle=0]{./imgs/supp/3d_rotation.pdf}
\end{center}
\vspace{-0.5cm}
\caption{\label{3d_rotate}\small{Rotate results on 3D faces by modify its boundary maps. There are $9$ blocks, the upper left block represents $5$ target boundary maps. For each blocks, the upper left face with red box is the input image and the rest $5$ are synthesized results corresponding to $5$ boundary maps in the first block. Each source has different texture and lighting.}}

\end{figure*}
\fi
\begin{figure*}[htb]
\begin{center}
\includegraphics[width=1\linewidth, angle=0]{./imgs/supp/pix_comp.pdf}
\end{center}
\vspace{-0.6cm}
\caption{\label{pix2pix}\small{Comparison with Pix2PixHD~\cite{Wang_2018_CVPR} on RaFD. The left image denotes the source image. The first row shows $8$ target boundary references. The second row shows images synthesized by pix2pixHD. The third row presents image generated by our algorithm. pix2pixHD takes boundary reference as input and thus are not able to keep source identity. }}
\vspace{-0.5cm}
\end{figure*}

\subsection{Comparison with Pix2pixHD}
we also conduct comparison with pix2pixHD~\cite{Wang_2018_CVPR}, a fully-supervised and general image-to-image translation framework which can also work well in our settings. The result can be found at Fig.~\ref{pix2pix}. As shown, pix2pixHD are likely to generate images with certain blur mainly due to its poor generalizability to unseen inputs. Besides, as it only takes boundary reference as its input, it is not able to keep the identity of source image, making generated results 'uncontrollable'. In contrast, our approach are able to synthesis photo-realistic faces with appealing perceptual quality as well as maintaining the source appearance. 

Besides, our model is also able to generate diverse outputs given fixed boundary reference input. As shown in Fig.~\ref{sample}, we sample appearance and condition on boundary maps. Compared to Pix2PixHD's ~`same' outputs when tested multiple times, our model has much higher diversity.

% Also, extensive face manipulation results on RafD dataset can be found at Fig.~\ref{rafd}.

\subsection{Face Manipulation on 3D Synthetic Data}

In order to better observe variations on appearance during manipulation, also expect light/texture-preserving nature of our model, we also conduct qualitative experiments on 3D synthetic face manipulation in Fig.~\ref{3d_mani}.

As shown, factors including skin,texture,lighting can be well preserved, which better testify to the great disentanglement of our model.

\subsection{Face Manipulation from single image}
In this subsection, we conduct face manipulation from a single image. By randomly modify the landmarks/boundary maps, our model are able to generate diverse and photo-realistic results which maintains the given appearance. Manipulating results can be found at Fig.~\ref{rafd}.

\subsection{Non-human Cases}
Experiments on non-human datasets are aimed at testifying the generalizability of our model. We conduct experiments on hands, cats dataset with the same hyper-parameter settings used in face datasets. Besides, we also test our CelebA-trained model on other stylized faces.

We test our model which trained using real world human face dataset(CelebA) on unreal stylized faces(e.g. cartoon, sketch) in Fig.~\ref{comic}, presenting the robustness of our model.

\begin{figure*}[htb]
\begin{center}
\includegraphics[width=1.0\linewidth, angle=0]{./imgs/supp/sample.pdf}
\end{center}
\vspace{-0.5cm}
\caption{\label{sample} \small{Comparison with pix2pixHD~\cite{Wang_2018_CVPR} on the diversity of sampling results.  Two source boundary maps are taken as input. The first row shows images synthesized by pix2pixHD, each source genrating 4 samples. The second and third rows present images generated by our model with randomly sampled $8$ appearances from the dataset each source. }}
\vspace{-0.5cm}
\end{figure*}

\subsection{Face Animation and Video Generation}

We also conduct experiments on face animation and video generation. \textit{\textbf{Our method can perform video face animation from a single image}}. Given a single source image and an target video with its landmarks(could be manipulated), by feeding those original landmarks and source image to our model frame by frame, the generated videos could be animated face results where the source face acts like the the face in the original video. Our framework can generate high-quality video as well as maintaining correlations on \textbf{Facial Action Consistency}, as shown in Fig.~\ref{au}. 

\subsection{Missing Identity Issues in Arbitrary Face Manipulation}

As discussed in main paper, arbitrary face manipulation potentially leads to identity changing issues. In this section, we provide a quantitative analysis about identity in Table.~\ref{verification}, using Verification accuracy to compare with input's neutral face. During experiment, The model is trained on CelebA dataset and tested on Rafd dataset. Each input face image is assigned with $8$ target boundary maps. We report the accuracy of both using same source landmarks and random selecting from the dataset. It illustrate that landmark information implicit covers partial identity information. Our model is capable of preserving identities when the landmarks belongs to same source. Note that there exists no prior knowledge about identity, e.g. pre-trained face recognition model, indicating that our framework could implicitly encode identity information beyond appearance.

\begin{table}[htb]
\begin{center}
\resizebox{0.8\columnwidth}{!}{%
\begin{tabular}{c|ccc} 
\Xhline{1.2pt}
\multirow{2}{*}{Input(\times 8)} & \multirow{2}{*}{Frontal~} & \multicolumn{2}{c}{Profile}                          \\ 
\cline{3-4}
                       & & $\pm45^{\circ}$ & $\pm90^{\circ}$  \\ 
\hline
Random Source & 29.3\% & 20.7\% & 10.1\% \\ 
\hline
Same Source & 95.8\% & 87.3\% & 61.2\%  \\
\Xhline{1.2pt}
\end{tabular}
}
\vspace{-0.3cm}
\end{center}
\caption{\label{verification} \small{Face verification results on Rafd dataset. Generator is trained on CelebA. Input column denotes the landmark source from the original person or random. }}
\vspace{-0.3cm}
\end{table}

\begin{figure}[htb]
\begin{center}
\includegraphics[width=1\linewidth, angle=0]{./imgs/supp/comic.pdf}
\end{center}
\vspace{-0.6cm}
\caption{\label{comic}Testing results on special stylized face(cartoon, sketch). Note that the model is trained on real world CelebA dataset and haven't seen peculiar faces before. The first line on the left presents four source faces. For each image given, four manipulated results are provided to the right. }
\end{figure}

\begin{figure}[h]
\begin{center}
\includegraphics[width=0.9\linewidth, angle=0]{./imgs/supp/3d_man.pdf}
\end{center}
\vspace{-0.6cm}
\caption{\label{3d_mani}\small{Face manipulation results on 3D image by modify the location of its landmarks. Images on the left with blue boxes are source images.  Images on the right each odd row represents target boundary maps and each even row shows corresponding synthesized results.}}
\end{figure}

\begin{figure*}[htb]
\begin{center}
\includegraphics[width=1\linewidth, angle=0]{./imgs/supp/rafd.pdf}
\end{center}
\vspace{-0.5cm}
\caption{\label{rafd} \small{Extensive results about face manipulation on RafD dataset. For each row, the left most one is the input source face. Then the model is fed with the person's boundary maps of $8$ target expressions. On the right are $8$ synthesized results under target poses. Note that boundary maps each row used are different and unseen during training.}}
\end{figure*}

% \begin{figure*}[h]
% \begin{center}
% \includegraphics[width=1\linewidth, angle=0]{./imgs/supp/single_mani.png}
% \end{center}
% \vspace{-0.5cm}
% \caption{\label{single}\small{Manipulation results on single image by randomly modify the location of its landmarks. The model is trained on CelebA and tested on \textbf{EmotionNet}. The image on the left of each block is the source image. The rest $20$ images per-sample on the right are the manipulated results.}}
% \end{figure*}

\vspace{-0.5cm}
\begin{figure*}[t!]
\begin{center}
\includegraphics[width=0.9\linewidth, angle=0]{./imgs/supp/au.pdf}
\end{center}
\vspace{-1.1cm}
\caption{\label{au}\small{Analysis on \textit{Facial Action Consistency}. Given a \textit{single} image and a video, the output video is generated by the given image and landmarks in the video frame by frame. We use a facial action detector to obtain responses from our model. $8$ representative AUs are selected to show the correlation of AUs response between our result and the source video. Each graph shows AUs response vs time. }}
\end{figure*}

{\small
\bibliographystyle{ieee_fullname}
\bibliography{egbib}
}